\documentclass{article}

\PassOptionsToPackage{numbers, sort&compress}{natbib}
 \usepackage[preprint]{neurips_2026}

\usepackage{ragged2e} 
\usepackage[utf8]{inputenc} % allow utf-8 input
\usepackage[T1]{fontenc}    % use 8-bit T1 fonts
\usepackage{url}            % simple URL typesetting
\usepackage{booktabs}       % professional-quality tables
\usepackage{amsfonts}       % blackboard math symbols
\usepackage{nicefrac}       % compact symbols for 1/2, etc.
\usepackage{microtype}      % microtypography
\usepackage{amsmath,amsfonts,bm}

\def\eqref#1{equation~\ref{#1}}
\def\1{\bm{1}}

\DeclareMathAlphabet{\mathsfit}{\encodingdefault}{\sfdefault}{m}{sl}
\SetMathAlphabet{\mathsfit}{bold}{\encodingdefault}{\sfdefault}{bx}{n}

\usepackage{amsmath}
\usepackage{amssymb}
\usepackage{mathtools}
\usepackage{amsthm}
\usepackage{algpseudocode}
\usepackage{algorithm}
\usepackage[table]{xcolor}
\usepackage{makecell}
\usepackage[most]{tcolorbox}

\usepackage{url}            % simple URL typesetting
\usepackage{booktabs}       % professional-quality tables
\usepackage{amsfonts}       % blackboard math symbols
\usepackage{nicefrac}       % compact symbols for 1/2, etc.
\usepackage{microtype}      % microtypography
\usepackage{xcolor}         % colors
\usepackage{wrapfig}

\usepackage{microtype}
\usepackage{graphicx}
\usepackage{booktabs} % for professional tables
\usepackage{multirow}
\usepackage{enumitem}
\usepackage{subcaption}
\usepackage{caption}
\definecolor{cvprblue}{rgb}{0.21,0.49,0.74}
\usepackage{hyperref}
\hypersetup{
    colorlinks=true,
    linkcolor=black,    
    urlcolor=cvprblue,
    citecolor=black
    }

\newcommand{\dr}{DeltaRubric~}

\title{DeltaRubric: Generative Multimodal Reward Modeling via Joint Planning and Verification}

\author{
Rui Liu$^{1,2}$,
Dian Yu$^{1}$,
Zhenwen Liang$^{1}$,
Yucheng Shi$^{1}$,
Tong Zheng$^{2}$,
Runpeng Dai$^{3}$, \\ 
\textbf{Haitao Mi}$^{1}$,
\textbf{Pratap Tokekar}$^{2}$,
\textbf{Leoweiliang}$^{1}$ \\
\\ 
$^{1}$Tencent Hunyuan
$^{2}$University of Maryland, College Park \\
$^{3}$University of North Carolina, Chapel Hill \\
Website: \href{https://deltarubric.github.io} {https://deltarubric.github.io}
}

\begin{document}

\maketitle

\vspace{-15pt}
\begin{abstract}
\vspace{-10pt}
Aligning Multimodal Large Language Models (MLLMs) requires reliable reward models, yet existing single-step evaluators can suffer from lazy judging, exploiting language priors over fine-grained visual verification. While rubric-based evaluation mitigates these biases in text-only settings, extending it to multimodal tasks is bottlenecked by the complexity of visual reasoning. The critical differences between responses often depend on instance-specific visual details. Robust evaluation requires dynamically synthesizing rubrics that isolate spatial and factual discrepancies. To address this, we introduce \textbf{DeltaRubric}, an approach that reformulates multimodal preference evaluation as a plan-and-execute process within a single MLLM. \dr operates in two steps: acting first as a \textit{Disagreement Planner}, the model generates a neutral, instance-specific verification checklist. Transitioning into a \textit{Checklist Verifier}, it executes these self-generated checks against the image and question to produce the final grounded judgment. We formulate \dr as a multi-role reinforcement learning problem, jointly optimizing planning and verification capabilities. Validated on Qwen3-VL 4B and 8B Instruct models, \dr achieves solid empirical gains. For instance, On VL-RewardBench, it improves base model overall accuracy by \textbf{+22.6} (4B) and \textbf{+18.8} (8B) points, largely outperforming standard no-rubric baselines. The results demonstrate that decomposing evaluation into structured, verifiable steps leads to more reliable and generalizable multimodal reward modeling.

% Aligning Multimodal Large Language Models (MLLMs) requires reliable reward models, yet existing single-step approaches often suffer from lazy judging, exploiting language priors rather than performing fine-grained visual verification. While rubric-based evaluation mitigates these biases in text-only tasks, multimodal evaluation demands dynamic, instance-specific criteria to resolve unique visual contradictions. To address this, we introduce \textbf{DeltaRubric}, a framework that reframes preference evaluation as an active, disagreement-driven visual investigation. Operating within a single shared MLLM, DeltaRubric acts first as a disagreement planner to isolate factual divergences between candidates into an actionable verification checklist, and subsequently as a checklist verifier to rigorously execute this rubric against the visual evidence. We jointly optimize these interdependent capabilities using a multi-role reinforcement learning framework.

\end{abstract}
% evidence check
% Self-Directed Evaluation via Disagreement Rubrics (SDER), reframing preference evaluation into an active visual investigation

% \input{intro}
\vspace{-10pt}
\section{Introduction} \label{sec: intro}
\vspace{-6pt}

Reinforcement Learning from Human Feedback (RLHF) \citep{ouyang2022training, bai2022training} has become the de facto standard for aligning Large Language Models (LLMs) with human intentions and values. At its core lies the Reward Model (RM) \citep{zhong2025comprehensive, lambert2025reinforcement, rafailov2023direct}, which serves as a proxy for human preference by scoring or comparing candidate responses and guiding policy optimization. For easy-to-verify tasks such as mathematical reasoning and coding, alignment can often be achieved with rule-based verifiers~\cite{trung-etal-2024-reft,lambert2024tulu,guo2025deepseek, zheng2025parallel, dai2025cde, liu2025vogue, liu2025stable, li2025self}. In contrast, open-ended and hard-to-verify tasks rely on learned reward models, which demand extensive human annotations to approximate nuanced preferences.

Recent advances have sought to move beyond scalar reward signals. In text-only settings, reward modeling has evolved from predicting scalar scores \citep{ouyang2022training, bai2022training} to LLM-as-a-judge frameworks \citep{saunders2022self, zheng2023judging, kim2023prometheus}, which generate both preference judgments and Chain-of-Thought (CoT) rationales \citep{wei2022chain}. To better capture the multidimensional nature of response quality in open-ended tasks, there is a growing trend toward adopting rubric-based evaluation~\citep{gunjal2025rubrics, liu2025openrubrics, huang2025reinforcement, shao2025dr, xu2026alternating}, including the most recent DeepSeek-V4~\citep{deepseekai2026deepseekv4}, demonstrating that decomposing a complex judgment into a set of criteria effectively improves evaluator reliability and generalization.

The transition toward Multimodal Large Language Models (MLLMs) introduces new alignment challenges~\citep{sun2024aligning, yu2024rlhf, chen2024mllm}. Extending RLHF to the visual domain requires multimodal reward models capable of assessing the consistency between textual claims and visual evidence. Existing multimodal reward models largely adopt a single-step paradigm, directly mapping inputs to a holistic preference or rationale. However, this single-step evaluation can suffer from lazy judging, a phenomenon where models bypass the demanding task of fine-grained decisions \citep{zheng2023judging}. Instead, they exploit textual priors or length biases, failing to rigorously verify the response against the image context \citep{singhal2023long, huang2024opera}. Furthermore, such evaluation often fails to capture the multifaceted nature of response quality, especially in non-verifiable domains \citep{xu2026alternating}. We argue that this formulation is limited: multimodal evaluation should not be treated as a passive scoring task, but rather an active reasoning process.

Although rubric-based evaluation has proven effective at mitigating these issues in text-only tasks, it remains largely underexplored in the multimodal domain. The primary bottleneck is the complexity of visual reasoning: the critical differences between two multimodal responses often depend on highly specific, instance-level visual details, such as object counts, spatial relationships, or localized hallucinations \citep{yu2024rlhf}. Consequently, multimodal evaluation demands an active mechanism capable of dynamically synthesizing instance-specific rubrics that isolate the critical factual and spatial discrepancies between responses. This limitation leads to a crucial research question: \textit{How can we transform multimodal preference evaluation from a single-step, holistic judgment into a structured, disagreement-driven verification process?}

\begin{figure}
    \centering
    \includegraphics[width=0.9\linewidth]{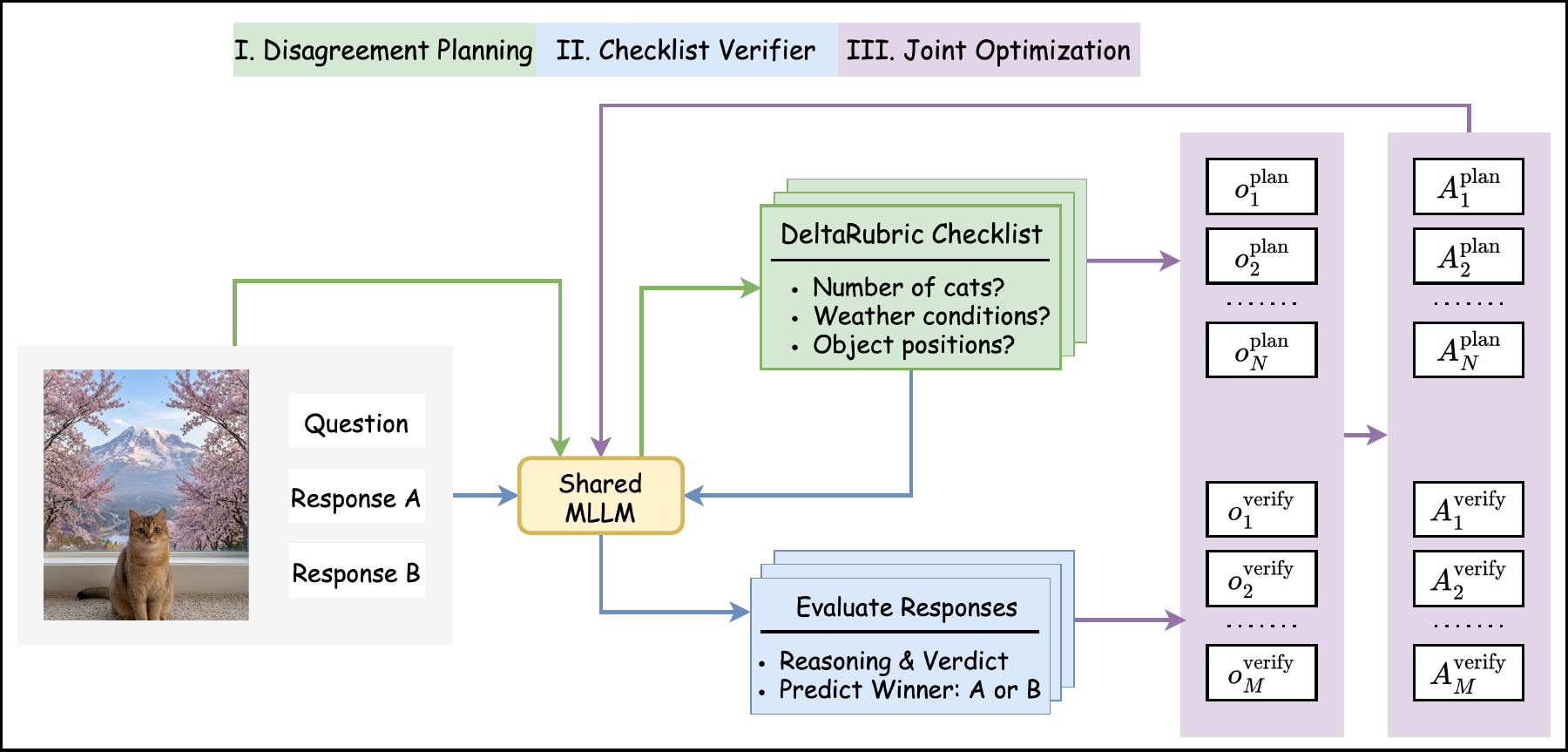}
    \caption{\textbf{Overview of DeltaRubric.} Given an input tuple $(I, q, y_A, y_B)$, a single shared MLLM operates in two sequential roles. As a \textit{Disagreement Planner}, the model analyzes the candidate responses to identify critical factual divergences and generates a neutral, instance-specific verification checklist. Example generated checklists are shown in Figure~\ref{fig:case1} and Appendix~\ref{app:examples}. Conditioned on this self-generated checklist, the same model then acts as a \textit{Checklist Verifier}, executing each item against the image $I$ and question $q$ to produce a grounded final judgment. We formulate \dr as a multi-role RL problem where planning and verification are optimized jointly.}
    \label{fig:app}
    \vspace{-15pt}
\end{figure}

To answer this, we introduce \textbf{DeltaRubric}, a framework that structurally decomposes multimodal evaluation into a sequential, two-step process within a single shared MLLM, as illustrated in Figure \ref{fig:app}. Rather than mapping multimodal inputs directly to a verdict, we reformulate evaluation as a plan-and-execute procedure, where the model first induces an explicit verification structure and then executes it for judgment. First, acting as a \textit{Disagreement Planner}, the model analyzes two candidate responses to isolate critical factual divergences, generating a neutral, actionable, and instance-specific verification checklist. Second, transitioning into a \textit{Checklist Verifier}, the model executes each item on the checklist against the visual evidence, aggregating the grounded findings to reach a final judgment.

% with role-specific objectives: the Planner is rewarded for producing checklists that improve downstream verification accuracy, while the Verifier is rewarded for correct, evidence-grounded decisions

% https://app.diagrams.net/?src=about#G1A-4pL8M_98K36aUZJ_uKIRHgvNQI3yA9#%7B%22pageId%22%3A%22M1vpOJiTVCOFTzo6cebW%22%7D

Training a single model to perform both planning and verification introduces a key challenge: how to jointly optimize planning quality and verification accuracy? To address this, we formulate \dr as a multi-role reinforcement learning problem, where planning and verification are optimized with distinct yet coordinated objectives. The Planner is rewarded for generating rubric checklists that expose and correct the Verifier’s blind spots, while the Verifier is rewarded for accurate and grounded execution. Inspired by recent generative reward modeling paradigms \citep{xu2026alternating, shao2025dr}, we move beyond static scalar rewards and instead optimize the model’s evaluative reasoning process itself. Using group-based RL algorithms such as GRPO \citep{shao2024deepseekmath} and DAPO~\cite{yu2025dapo}, we compute task-specific advantages and update both capabilities through a shared policy. This design enables the model to internalize evaluation as a structured, verification-driven reasoning process, resulting in a robust generative reward model that generalizes across complex multimodal tasks.

We validate our approach by training Qwen3-VL 4B and 8B Instruct \citep{bai2025qwen3} models and evaluating them on a comprehensive benchmark suite. On VL-RewardBench \citep{li2025vl}, \dr improves overall accuracy of base models by +22.6 (4B) and +18.8 (8B) points, and consistently outperforms the no-rubric baselines (+4.3 and +8.1, respectively). On Multimodal RewardBench \citep{yasunaga2025multimodal}, it improves the overall accuracy of the 8B base model by +5.5 and surpasses the no-rubric baseline by +4.5. Furthermore, we evaluate on the text-only RewardBench \citep{lambert2025rewardbench}, \dr elevates the 8B base model's overall accuracy by +3.2, indicating that multimodal finetuning with \dr preserves, and even enhances foundational language capabilities. Overall, these results suggest that decomposing evaluation into structured, verifiable steps leads to more reliable and generalizable multimodal reward modeling. In summary, our work offers the following key contributions: 

\vspace{-6pt}
\begin{itemize}[left=0pt]
    \item We propose DeltaRubric, a novel approach that reframes multimodal evaluation as an active, two-step visual investigation.
    \item By decoupling the evaluation process into a Planner and Verifier, optimized jointly via multi-role RL, \dr encourages the model to isolate factual contradictions and ground its judgments in visual evidence, effectively mitigating lazy judging and improving evaluation reliability.
    \item \dr achieves solid empirical gains. On VL-RewardBench, it improves base model overall accuracy by \textbf{+22.6} (4B) and \textbf{+18.8} (8B) points, largely outperforming standard no-rubric baselines. Furthermore, text-only RewardBench evaluations demonstrate that \dr prevents catastrophic forgetting while actively enhancing foundational structural logic.
\end{itemize}

\vspace{-8pt}
\section{Related Work} \label{sec: related}

\vspace{-6pt}
\paragraph{Multimodal Reward Modeling.}
The alignment of MLLMs heavily relies on extending RLHF to the visual domain, necessitating robust multimodal reward models \citep{sun2024aligning, yu2024rlhf}. Early efforts primarily adapted the LLM-as-a-judge paradigm \citep{zheng2023judging, chen2024mllm} to evaluate the interaction between textual claims and visual inputs \citep{xiong2025llava, yu2025rlaif}. Recently, progress has been made in both optimizing direct scalar reward baselines \citep{zhang2026basereward} and developing generative multimodal reward models that incorporate CoT reasoning to improve reliability \citep{wang2025unified, zhang2025r1, wang2026msrl}. Regardless of the specific architecture, most methods train a monolithic model to process visual inputs and output either a direct preference score or a holistic rationale \citep{zhang2025mm}.

Despite these advancements, monolithic multimodal evaluators exhibit biases similar to the lazy judging phenomenon observed in text-based LLMs \citep{zheng2023judging}. Because fine-grained visual grounding is inherently challenging, models often bypass rigorous image verification and instead exploit language priors, formatting, or length biases \citep{singhal2023long, huang2024opera}. While recent work has attempted to reinforce visual reasoning via agentic tool use \citep{ding2025arm}, current methods still lack an intrinsic mechanism that enforces explicit visual investigation. Our framework, DeltaRubric, bridges this gap by shifting evaluation from a passive scoring task to an active, two-step process. By structurally decoupling the isolation of contested textual claims (the Planner) from the grounded verification against visual evidence (the Verifier), DeltaRubric neutralizes textual bias and enforces structured, instance-level visual verification.

\vspace{-10pt}
\noindent \paragraph{Rubrics as Rewards.}
To address the opacity and unreliability of direct preference scoring in open-ended, non-verifiable tasks, the text-only domain has increasingly adopted rubric-based and checklist-driven evaluation frameworks \citep{viswanathan2025checklists, gunjal2025rubrics}. By decomposing complex judgments into explicitly defined criteria, these methods reduce cognitive load and improve reward model alignment \citep{huang2025reinforcement, liu2025openrubrics}. Recent approaches have further scaled these concepts, utilizing alternating reinforcement learning and self-evolving rubrics to reinforce CoT reasoning, handle deep research, and guide non-verifiable post-training \citep{sheng2026reinforcing, shao2025dr, xu2026alternating}.

However, the application of rubric-based rewards to the multimodal domain remains largely underexplored. Unlike text evaluation, visual evaluation requires verifying highly specific, instance-level physical realities, such as localized hallucinations, object counts, and spatial relationships \citep{yu2024rlhf}. While recent work has begun exploring rubric-based generative rewards for multimodal reasoning \citep{jia2025autorubric}, existing pipelines rely on disjointed architectures prone to cascading errors. Our approach addresses this by dynamically synthesizing disagreement-focused rubrics directly from candidate conflicts. Furthermore, unlike previous approaches that rely on separate models for rubric generation and preference evaluation \citep{xu2026alternating}, DeltaRubric jointly optimizes both capabilities via multi-role reinforcement learning. The decoupled advantage estimation ensures the model learns to actively hunt for critical visual discrepancies without cross-task variance corrupting the learning signal.

\vspace{-6pt}
\section{Approach}
\label{sec: approach}
\vspace{-6pt}

We present \textbf{DeltaRubric}, a framework for multimodal reward modeling that decomposes evaluation into a self-guided, two-step process within a single shared MLLM. Instead of directly predicting a scalar reward or binary verdict from a visual prompt, \dr first generates a disagreement-focused verification checklist (the \textit{Planner}) and then executes this checklist against the image to derive the final judgment (the \textit{Verifier}). Both roles are jointly optimized through multi-role RL.

\vspace{-6pt}
\subsection{Problem Formulation}
\vspace{-6pt}

Let each training sample be denoted as $x = (I, q, y_\text{A}, y_\text{B})$, where \(I\) is the image, \(q\) is the question, and $y_\text{A}, y_\text{B}$ are two candidate responses, with $z^* \in \{\text{A}, \text{B}\}$ representing the preferred response. The objective is to predict the superior response $z \in \{\text{A}, \text{B}\}$. Standard RLHF approaches directly model $\pi_\theta(z \mid x)$ or generate reasoning before prediction $\pi_\theta(r, z \mid x)$. However, such single-step evaluation is prone to lazy judging, where models rely on textual priors or superficial patterns instead of grounded visual verification. To address this, we reformulate multimodal evaluation as a latent plan-generation and execution problem mediated by an intermediate, self-generated verification checklist $c$.

\vspace{-6pt}
\subsection{Planner-Verifier Architecture}
\vspace{-6pt}
To enforce fine-grained and grounded evaluations, we introduce a shared policy model $\pi_\theta$ that acts consecutively as a \textit{Planner} and a \textit{Verifier}. 

\vspace{-6pt}
\paragraph{Disagreement Planner.} Given an input tuple $x = (I, q, y_A, y_B)$, the Planner generates a checklist $c \sim \pi_\theta(\cdot \mid x)$. The checklist consists of a short sequence of verifiable constraints (e.g., concrete visual attributes, object counts, spatial relations, or hallucinated claims) identifying exactly where the two candidate responses fundamentally disagree. We prompt the model to output a strictly neutral, evidence-seeking checklist without expressing a preference for either candidate. A post-generation filtering step is applied to further enforces this impartiality. The generated checklist examples can be seen in Figure \ref{fig:case1} and Appendix \ref{app:examples}. The prompt template for generating the checklist is provided in Appendix \ref{app:prompt}. 

\vspace{-6pt}
\paragraph{Checklist Verifier.} The Verifier takes the original input along with the generated checklist to produce the final evaluation. It generates a step-by-step reasoning trajectory $r$ followed by the final verdict $z$: $(r, z) \sim \pi_\theta(\cdot \mid x, c)$. The Verifier explicitly evaluates each item on the checklist against the image $I$ before aggregating the evidence to decide the winner. The Verifier is instructed to treat the checklist as a shortlist of checks to execute, ignoring any checks that are hallucinated or contradicted by the image. The prompt template for verifier evaluation can be found in Appendix \ref{app:prompt}.

\vspace{-6pt}
\subsection{Joint Optimization via Multi-Role RL}
\vspace{-6pt}
We jointly optimize the Planner and Verifier capabilities of the shared model $\pi_\theta$ through a multi-role reinforcement learning objective. During each training iteration, $\pi_\theta$ performs both tasks sequentially, generating distinct sets of rollouts for planning and verification. Crucially, the advantages for the Planner and the Verifier are computed independently within their respective task groups. This decoupled advantage estimation allows the isolated signals to be aggregated into a single, unified joint loss function for the final policy update, effectively preventing cross-task variance.

\vspace{-6pt}
\paragraph{Planner Learning.} For a given input $x$, we sample $N$ candidate checklists $\mathcal{C} = \{c_1, \dots, c_N\} \sim \pi_\theta(\cdot \mid x)$ from the current policy. To efficiently score each checklist $c_i$, we query the Verifier using a lightweight cheap probe prompt to obtain a fast verdict without extended reasoning: $z_i \sim \pi_\theta(z \mid x, c_i)$. The cheap probe prompt template is provided in  Appendix \ref{app:prompt}. Concurrently, we obtain a baseline verdict without providing any rubric checklist: $z_0 \sim \pi_\theta(z \mid x)$. The planner reward is defined by its relative ability to improve over the baseline accuracy:
\begin{equation} \label{eq: planner_reward}
    R_{plan}(c_i) = \mathbb{I}(z_i = z^*) - \mathbb{I}(z_0 = z^*),
\end{equation}
where $\mathbb{I}(\cdot)$ is the indicator function. Therefore, a checklist receives $+1$ reward if it flips an incorrect no-rubric baseline verdict to correct, $-1$ if it misleads the verifier into an error, and $0$ otherwise. We then calculate the Planner advantage $A_{plan}^{(i)}$ by normalizing the rewards within the group of $N$ candidate checklists:
$
    A_{plan}^{(i)} = \frac{R_{plan}(c_i) - \mu_{\mathcal{C}}}{\sigma_{\mathcal{C}}},
$
where $\mu_{\mathcal{C}}$ and $\sigma_{\mathcal{C}}$ are the mean and standard deviation of $\{R_{plan}(c_1), \dots, R_{plan}(c_N)\}$.

% We enforce syntax validity via a heuristic filter.

\vspace{-6pt}
\paragraph{Verifier Learning.} After scoring the Planner candidates, we run a greedy forward pass through the Planner: $c^* = \arg\max \pi_\theta(\cdot \mid x)$ and passes this greedy checklist to the Verifier. We then sample $M$ full reasoning trajectories and verdicts $\mathcal{R} = \{(r_1, z_1), \dots, (r_M, z_M)\} \sim \pi_\theta(\cdot \mid x, c^\star)$ from the Verifier conditioned on $c^*$. The Verifier is rewarded based on final accuracy and a conditional guidance bonus: 
\begin{equation} \label{eq: verifer_reward}
R_{verify}(r_j, z_j) = \mathbb{I}(z_j = z^*)  + \lambda \max\left(0,
      \mathbb{I}(z_j = z^*) - \mathbb{I}(z_0 = z^*)\right).
\end{equation}

Here, the final accuracy term rewards correct verdicts, while the bonus term specifically rewards cases where checklist-guided verification strictly improves upon the no-guidance baseline, enforced by a $\max(0, \cdot)$ threshold. Similarly, the Verifier advantage $A_{verify}^{(j)}$ is normalized strictly within the $M$ verifier trajectories generated for that prompt:
$
   A_{verify}^{(j)} = \frac{R_{verify}(r_j, z_j) - \mu_{\mathcal{R}}}{\sigma_{\mathcal{R}}}.
$
\vspace{-6pt}
\paragraph{Joint Multi-Role Loss.} The final policy update combines the separate experiences into a single optimization step. Let $\mathcal{L}(\theta \mid \text{data}, A)$ denote the standard RL clipped surrogate objective (e.g., GRPO). The shared model $\theta$ is updated by minimizing the joint loss:
\begin{equation}
   \mathcal{L}_{total}(\theta) = \mathbb{E}_{x \sim \mathcal{D}} \left[ \frac{1}{N} \sum_{i=1}^N \mathcal{L}\left(\theta \mid
      c_i, A_{plan}^{(i)}\right) + \frac{1}{M} \sum_{j=1}^M \mathcal{L}\left(\theta \mid r_j, z_j, A_{verify}^{(j)}\right) \right].
\end{equation}

By computing advantages separately for each task group, we ensure that the Planner gradients are strictly driven by checklist quality, and Verifier gradients are strictly driven by execution quality, preventing cross-task variance from corrupting the RL signals.

\vspace{-6pt}
\section{Experiments} \label{sec: exp}
\vspace{-6pt}

\subsection{Experimental Setup}

\vspace{-6pt}
\paragraph{Implementation Details.}
We conduct direct RL fine-tuning on the Qwen3-VL-4B and 8B Instruct models \citep{bai2025qwen3}, utilizing GRPO as the underlying RL algorithm. During training, we sample $\text{N}=5$ candidate checklists per prompt for the Planner, and $\text{M}=5$ reasoning trajectories per prompt for the Verifier. For the verifier reward defined in Eq. \ref{eq: verifer_reward}, we set the guidance bonus coefficient to $\lambda=0.4$; please see a justification for this value via a sensitivity analysis in Appendix \ref{app:sen}. Our implementation is built on the EasyR1 framework \citep{zheng2025easyr1}. More details are provided in Appendix \ref{app:imp}.

\vspace{-6pt}
\paragraph{Dataset and Benchmarks.}
To construct the training dataset, we randomly sample 30K instances from the RLAIF-V dataset \citep{yu2025rlaif}.  Each instance consists of an image-query pair, two candidate responses, and a preference label. We strictly decontaminate this data to ensure zero overlap with our evaluation sets. We validate our approach on rigorous benchmarks: VL-RewardBench \citep{li2025vl}, an out-of-domain set designed to probe robustness to common failure modes such as visual hallucinations and spatial reasoning errors; and Multimodal RewardBench \citep{yasunaga2025multimodal}, which evaluates general vision-language preference alignment.

% We explicitly remove any samples that overlap with VL-RewardBench \citep{li2025vl} to ensure zero contamination with evaluation benchmarks.

\vspace{-6pt}
\paragraph{Baselines.}
For our controlled baseline comparisons, we evaluate: (1) a zero-shot base model, where off-the-shelf models are prompted to act as a judge without any RL fine-tuning; (2) a no-rubric setting, where RL-finetuned models generate a CoT rationale followed by a verdict, representing a standard reward modeling paradigm. In addition, for broader context, we include evaluated results from several external models, including SliME \citep{zhang2024benchmarking}, VITA-1.5 \citep{fu2025vita}, Molmo-7B \citep{deitke2025molmo}, InternVL2/3-8B \citep{chen2024internvl, zhu2025internvl3}, Llama-3.2 \citep{grattafiori2024llama}, Molmo-7B \citep{deitke2025molmo}, MM-RLHF-Reward-7B \citep{zhang2025mm}, LLaVA-Critic-8B \citep{xiong2025llava}, and NVLM-D-72B \citep{dai2024nvlm}.

\vspace{-6pt}
\subsection{Main Results}
\begin{figure*}[t]
    \centering
    % Subfigure 1
    \begin{subfigure}[t]{0.325\linewidth} % Allocate ~1/3 width
        \centering
        \includegraphics[width=\textwidth]{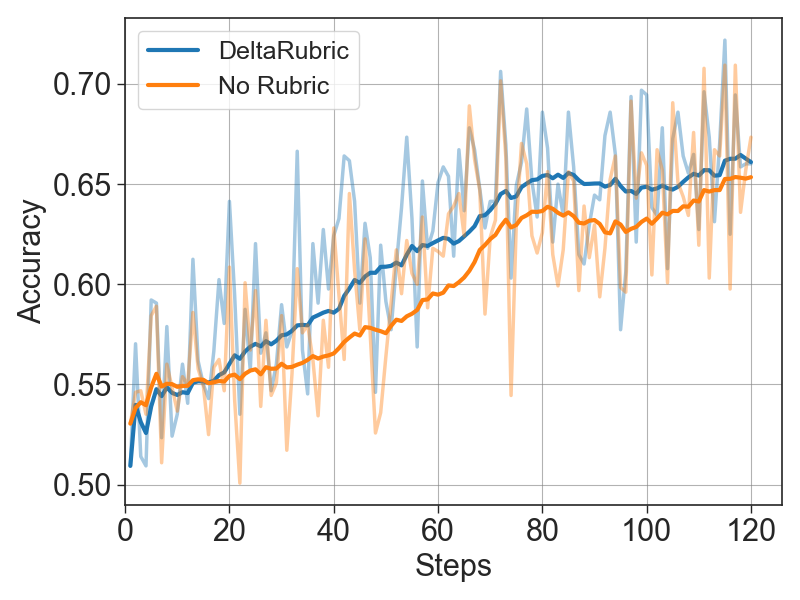}
        \caption{Verifier training.}
        \label{fig:verifer_acc}
    \end{subfigure}
    % Subfigure 2
    \begin{subfigure}[t]{0.325\linewidth}
        \centering
        \includegraphics[width=\textwidth]{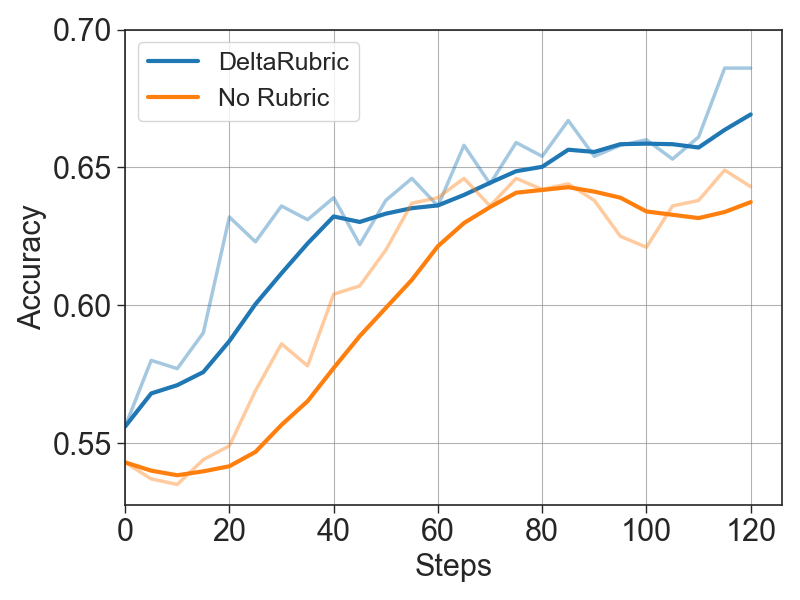}
        \caption{Verifier validation.}
        \label{fig:val_acc}
    \end{subfigure}
    % Subfigure 3
    \begin{subfigure}[t]{0.325\linewidth}
        \centering
        \includegraphics[width=\textwidth]{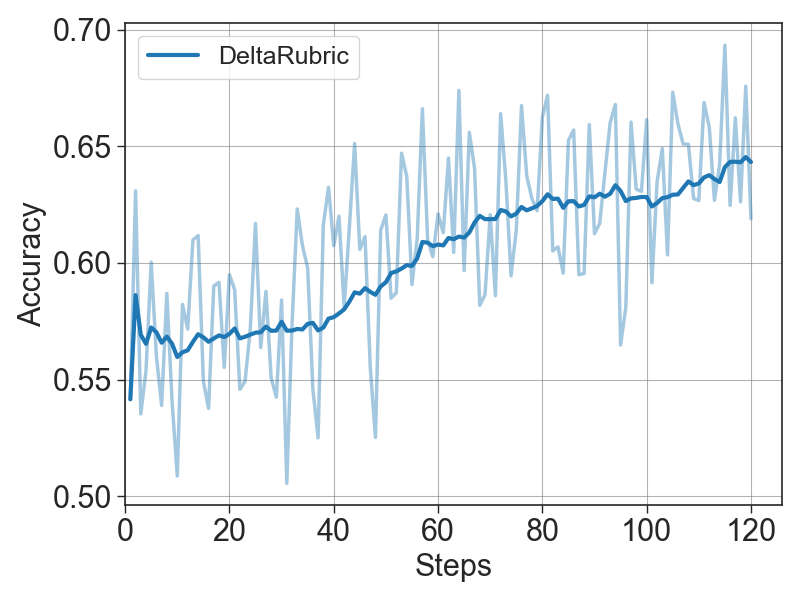}
        \caption{Planner probe accuracy.}
        \label{fig:planner_acc}
    \end{subfigure}
    \caption{\textbf{Planner and Verifier training dynamics.} \textbf{(a)} and \textbf{(b)} compare the Verifier's training and validation accuracy (evaluated every 5 steps) against a no-rubric baseline. While both approaches improve, \dr overall achieves a higher mean accuracy in juding final responses. \textbf{(c)} tracks the Planner probe accuracy, an intermediate proxy measuring the fraction of sampled checklists that successfully guide a lightweight verdict probe to the correct ground-truth winner. The steady upward trajectory confirms the Planner is learning to generate increasingly decision-useful rubric checklists.}
    \label{fig:train}
    \vspace{-8pt}
\end{figure*}

\vspace{-6pt}
We illustrate the training dynamics of the Planner and Verifier in Figure \ref{fig:train}. Figure \ref{fig:verifer_acc} compares the Verifier training accuracy of \dr against the no-rubric baseline. While both approaches improve over time, \dr achieves higher accuracy in evaluating the final responses. This trend is further supported by the validation accuracy (measured every five steps) in Figure \ref{fig:val_acc}. Additionally, Figure \ref{fig:planner_acc} plots the Planner probe accuracy, measured as the fraction of sampled checklists that successfully guide a lightweight verdict probe to the correct ground-truth winner. This metric serves as a proxy for checklist quality. Its steady increase indicates that the generated checklists become progressively more decision-useful over training, enabling the probe to make more accurate judgments. This improvement aligns with the gains observed in Verifier performance (Figures~\ref{fig:verifer_acc} and \ref{fig:val_acc}), highlighting the effectiveness of \dr.

We then present the evaluation results on VL-RewardBench in Table \ref{tab:res_vl}. Following the evaluation protocol of \citep{li2025vl}, we compute accuracy via greedy decoding. We report subcategory accuracy (the proportion of correct predictions within each subset), overall accuracy (performance across the entire dataset), and the macro-average (the mean of all subcategory accuracies). As shown, while applying standard preference optimization without rubrics improves upon the base capabilities of both the Qwen3-VL 4B and 8B Instruct models, \dr drives larger gains. Specifically, our framework outperforms the no-rubric baseline in overall accuracy by 4.3 and 8.1 points for the 4B and 8B models.

\begin{table*}[t]
    \centering
    \caption{\textbf{Evaluation on the VL-RewardBench.} \dr improves the overall accuracy of the Qwen3-VL-4B and 8B Instruct base models by \textbf{+22.6} and \textbf{+18.8} points, respectively. Crucially, it consistently outperforms the standard no-rubric baseline across both architectures (\textbf{+4.3} points for the 4B model and \textbf{+8.1} points for the 8B model), demonstrating the effectiveness of our approach.}
    \vspace{-3pt}
    \resizebox{\textwidth}{!}{
    \begin{tabular}{lccccc}
        \toprule
        \textbf{Models} & \textbf{General} & \textbf{Hallucination} & \textbf{Reasoning} & \textbf{Overall} & \textbf{Macro Avg} \\
        % \midrule
        % \multicolumn{6}{c}{\textit{Proprietary Models}} \\
        % \midrule
        % % Gemini-1.5-Flash (2024-09-24) & 47.8 & 59.6 & 58.4 & 57.6 & 55.3 \\
        % GPT-4o \citep{hurst2024gpt} & 49.1 & 67.6 & 70.5 & 65.8 & 62.4 \\
        % Gemini-1.5-Pro \citep{team2024gemini} & 50.8 & 72.5 & 64.2 & 67.2 & 62.5 \\
        % Claude-3.5-Sonnet \citep{claude35} & 43.4 & 55.0 & 62.3 & 55.3 & 53.6 \\
        % GPT-4o-mini (2024-07-18) & 41.7 & 34.5 & 58.2 & 41.5 & 44.8 \\
        
        \midrule
        \multicolumn{6}{c}{\textit{Open-Source Models}} \\
        \midrule
        VITA-1.5-7B \citep{fu2025vita} & 18.6 & 8.9 & 22.1 & 16.5 & 16.5 \\
        SliME-7B \citep{zhang2024benchmarking} & 7.2 & 27.1 & 18.6 & 19.0 & 17.6 \\
        Molmo-7B \citep{deitke2025molmo} & 31.1 & 31.8 & 56.2 & 37.5 & 39.7 \\
        MM-RLHF-Reward-7B \citep{zhang2025mm} & 45.0 & 50.5 & 57.6 & 50.2 & 51.0 \\ 
        InternVL2-8B \citep{chen2024internvl} & 35.6 & 41.1 & 59.0 & 44.5 & 45.2 \\
        LLaVA-Critic-8B \citep{xiong2025llava} & 54.6 & 38.3 & 59.1 & 44.5 & 45.2 \\
        Llama-3.2-11B \citep{grattafiori2024llama} & 33.3 & 38.4 & 56.6 & 42.9 & 42.8 \\
        NVLM-D-72B \citep{dai2024nvlm} & 38.9 & 31.6 & 62.0 & 40.1 & 44.1 \\
        Llama-3.2-90B \citep{grattafiori2024llama} & 42.6 & 57.3 & 61.7 & 56.2 & 53.9 \\
        % UnifiedReward-7B \citep{wang2025unified} & 43.6 & 65.7 & 64.4 & 62.1 & 57.9 \\
        % R1-Reward \citep{zhang2025r1} & 54.6 & 77.7 & 48.7 & 62.2 & 60.3 \\

        \midrule
        \multicolumn{6}{c}{\textit{DeltaRubric}} \\
        \midrule
        Qwen3-VL-4B Instruct \citep{bai2025qwen3} & 46.4 & 64.9 & 36.0 & 54.9 & 49.1 \\
        \quad + No rubric & 51.9 & 87.1 & 50.8 & 73.2 & 63.3 \\
        % \quad + Static rubric & 47.5 & 85.7 & 62.8 & 74.3 & 65.3 \\
        \quad + \dr & 55.3 & 87.7 & 65.9 & 77.5 & 69.6 \\
        
        \midrule
        Qwen3-VL-8B Instruct \citep{bai2025qwen3} & 47.0 & 72.4 & 43.2 & 61.3 & 54.2 \\
        \quad + No rubric & 55.8 & 86.1 & 48.3 & 72.0 & 63.4 \\
        % \quad + Static rubric & 53.6 & 87.2 & 59.6 & 75.3 & 66.8 \\
        \quad + \textbf{\dr} & \textbf{59.7} & \textbf{88.3} & \textbf{72.6} & \textbf{80.1} & \textbf{73.5} \\
        \bottomrule
    \end{tabular}
    }
    \label{tab:res_vl}
    \vspace{-15pt}
\end{table*}

Consequently, our approach achieves the best performance across all evaluation aspects of the benchmark. By explicitly forcing the model to generate a targeted disagreement checklist prior to evaluation, \dr ensures faithful visual verification. The Planner successfully isolates the exact attributes that distinguish the two candidate responses, while the Verifier, trained to strictly execute this checklist against the image, grounds the final verdict in empirical evidence. This structural intervention results in accuracy gains, effectively mitigating the lazy judging problem.

\begin{table*}[t]
    \centering
    \caption{\textbf{Evaluation on the Multimodal RewardBench.} \dr improves the overall accuracy of the Qwen3-VL-8B Instruct base model by \textbf{+5.5} points. Crucially, it outperforms the standard no-rubric preference optimization baseline by \textbf{+4.5} points, demonstrating the effectiveness of our instance-specific approach to synthesizing targeted rubrics.}
    \vspace{-3pt}
    \resizebox{\textwidth}{!}{
    \begin{tabular}{l c cc c cc c c}
        \toprule
        \multirow{2}{*}{\textbf{Model}} & \multirow{2}{*}{\textbf{Overall}} & \multicolumn{2}{c}{\textbf{General}} & \multirow{2}{*}{\textbf{Knowledge}} & \multicolumn{2}{c}{\textbf{Reasoning}} & \multirow{2}{*}{\textbf{Safety}} & \multirow{2}{*}{\textbf{VQA}} \\
        \cmidrule(lr){3-4} \cmidrule(lr){6-7}
        & & Correctness & Preference & & Math & Coding & & \\
        % \midrule
        % \multicolumn{9}{c}{\textit{Proprietary Models}} \\
        % \midrule
        
        % GPT-4o \citep{hurst2024gpt} & 71.5 & 62.6 & 69.0 & 72.0 & 67.6 & 62.1 & 74.8 & 87.2 \\
        % Gemini-1.5-Pro \citep{team2024gemini} & 72.0 & 63.5 & 67.7 & 66.3 & 68.9 & 55.5 & 94.5 & 87.2 \\
        % Claude-3.5-Sonnet \citep{claude35} & 72.0 & 62.6 & 67.8 & 73.9 & 68.6 & 65.1 & 76.8 & 85.6 \\
        
        \midrule
        \multicolumn{9}{c}{\textit{Open-Source Models}} \\
        \midrule
        VITA-1.5-7B \citep{fu2025vita} & 53.6 & 55.6 & 54.3 & 52.5 & 51.9 & 52.8 & 58.1 & 50.0 \\
        Molmo-7B \citep{deitke2025molmo} & 52.9 & 56.8 & 59.4 & 54.6 & 50.7 & 53.4 & 34.8 & 60.3 \\
        MM-RLHF-Reward-7B \citep{zhang2025mm} & 67.1 & 61.7 & 67.5 & 54.3 & 58.4 & 57.9 & 92.9 & 76.8 \\
        SliME-8B \citep{zhang2024benchmarking} & 42.0  & 42.3 & 52.2 & 47.5 & 43.5 & 35.3 & 19.1 & 53.8 \\
        InternVL3-8B \citep{zhu2025internvl3} & 63.6 & 59.6 & 61.6 & 60.5 & 65.1 & 56.6 & 59.3 & 82.3 \\ 
        Llama-3.2-11B \citep{grattafiori2024llama} & 51.2 & 57.8 & 65.8 & 55.5 & 50.6 & 51.7 & 20.9 & 55.8 \\
        Llama-3.2-90B \citep{grattafiori2024llama} & 61.2 & 60.0 & 68.4 & 61.2 & 56.3 & 53.1 & 52.0 & 77.1 \\
        % Aria & 57.3 & 59.5 & 63.5 & 55.5 & 50.3 & 54.2 & 46.1 & 64.2 \\
        % Llava-1.5-13B & 48.9 & 53.3 & 55.2 & 50.5 & 53.5 & 49.3 & 20.1 & 51.8 \\
        % R1-Reward \citep{zhang2025r1} & 51.3 & 52.3 & 51.4 & 51.4 & 49.4 & 44.5 & 58.5 & 51.8 \\
        % UnifiedReward-7B \citep{wang2025unified} & 63.5 & 64.7 & 62.8 & 61.0 & 58.8 & 58.2 & 81.1 & 61.8 \\
        
        \midrule
        \multicolumn{9}{c}{\textit{DeltaRubric}} \\
        \midrule
        Qwen3-VL-4B Instruct \citep{bai2025qwen3} & 65.3 & 66.1 & 56.3 & 52.7 & 46.7 & 54.4 & 80.4 & 70.8  \\
        \quad + No rubric & 66.4 & 74.5 & 60.1 & 59.4 & 54.3 & 55.0 & 87.6 & 71.3 \\
        % \quad + Static rubric & 67.0 & 72.6 & 61.5 & 61.3 & 51.4 & 53.6 & 89.4 & 78.6 \\
        \quad + \dr & 69.1 & 73.7 & 65.4 & 60.0 & 58.0 & 52.0 & 91.2 & 80.8 \\
        
        \midrule
        Qwen3-VL-8B Instruct \citep{bai2025qwen3} & 67.7 & 68.9 & 61.5 & 56.2 & 64.6 & 49.6 & 82.6 & 71.4 \\
        \quad + No rubric & 68.7 & 75.0 & 62.7 & 56.7 & 64.0 & 51.5 & 91.5 & 77.0 \\
        % \quad + Static rubric & 71.0 & 76.2 & 63.8 & 68.2 & 67.7 & 52.2 & 90.2 & 81.4 \\
        \quad + \textbf{\dr} & \textbf{73.2} & \textbf{76.9} & \textbf{65.9} & \textbf{69.5} & \textbf{68.7} & \textbf{52.6} & \textbf{93.3} & \textbf{84.9} \\
        \bottomrule
    \end{tabular}
    }
    \label{tab:res_mm}
    \vspace{-13pt}
\end{table*}

We then evaluate on the Multimodal RewardBench. We report the overall and subcategory accuracy in Table \ref{tab:res_mm}, following the evaluation protocol of \citep{yasunaga2025multimodal}. The results across both Qwen3-VL-4B and 8B models demonstrate a clear benefit to our proposed approach. For the 8B model, standard direct preference optimization without rubrics achieves a baseline accuracy of 68.7. Our approach improves the overall accuracy to 73.2, validating that instance-specific and disagreement-focused rubrics help capture the complex visual nuances of multimodal prompts.

% Introducing a static rubric improves this to 71.0, confirming that explicit structural guidance aids the evaluation process. 

\begin{figure*}
    \centering
    \includegraphics[width=\linewidth]{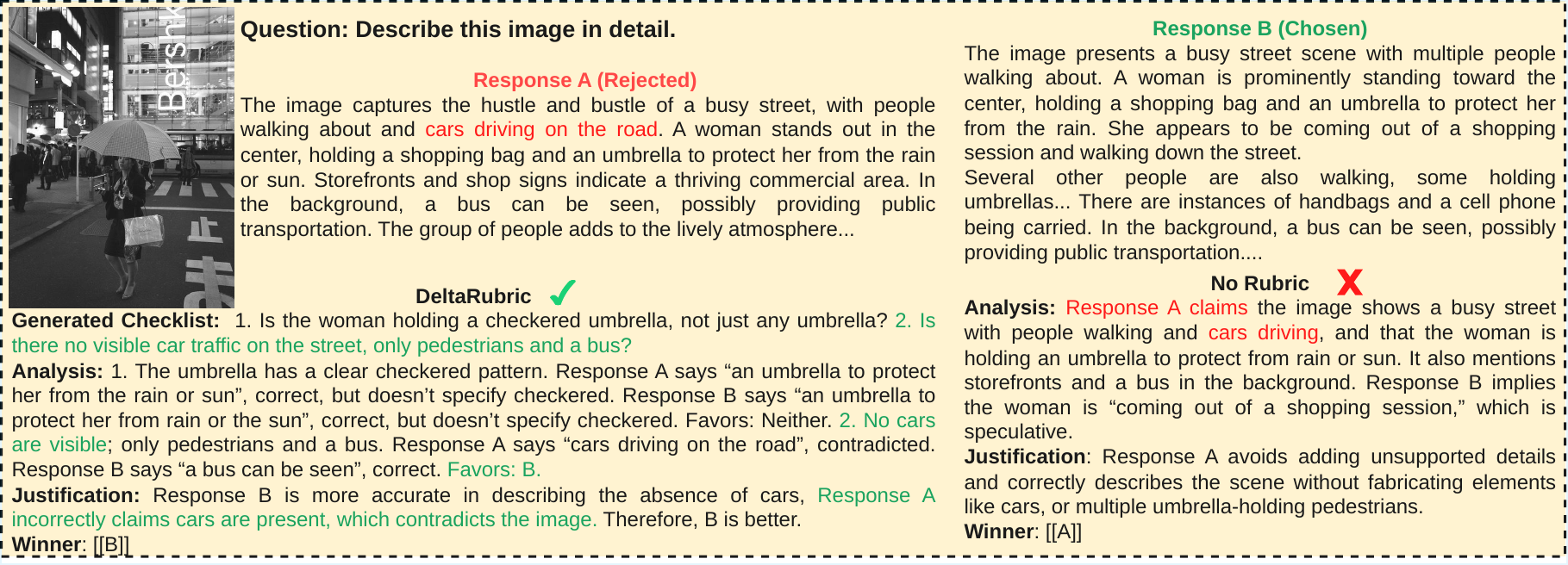}
    \caption{\textbf{Qualitative comparison of evaluation methods.} While the standard no-rubric baseline misses the hallucinated "cars" in Response A, \dr generates a targeted disagreement checklist. This explicitly enforces visual verification, allowing the model to successfully catch the hallucination and correctly select Response B.}
    \label{fig:case1}
    \vspace{-15pt}
\end{figure*}

Analyzing the subcategories reveals exactly where this active visual investigation excels. Our approach yields large gains in heavily visually-dependent tasks such as VQA and Safety. For the 8B model, VQA accuracy improves from a base of 71.4 to 84.9, while Safety accuracy reaches 93.3. This confirms that synthesizing targeted rubrics effectively neutralizes the lazy judging and textual priors that bottleneck traditional monolithic evaluation methods. We present a qualitative example in Figure \ref{fig:case1} comparing the evaluation outputs of the no-rubric baseline and \dr. As shown, while the standard no-rubric baseline misses the hallucinated "cars" in Response A, \dr generates a targeted disagreement checklist. This explicitly enforces visual verification, allowing the model to successfully catch the hallucination and correctly select Response B. More examples can be seen in Appendix \ref{app:examples}.

\begin{wraptable}{r}{0.5\textwidth}
    \vspace{-12pt}
    \centering
    \caption{\textbf{Ablation on Planner optimization.} Performance on VL-RewardBench. While a frozen, zero-shot Planner improves upon the base model, actively training the Planner via RL is beneficial, driving a \textbf{+6.3} point improvement in the Reasoning subcategory over its frozen counterpart.}
    \vspace{-5pt}
    \footnotesize 
    \setlength{\tabcolsep}{5pt} 
    \begin{tabular}{lccccc} 
        \toprule
        \textbf{Models} & \textbf{Gen.} & \textbf{Hallu.} & \textbf{Reas.} & \textbf{All} & \textbf{Macro} \\
        \midrule
        Base 8B & 47.0 & 72.4 & 43.2 & 61.3 & 54.2 \\
        Ours (Frozen) & 58.6 & \textbf{89.0} & 66.3 & 78.8 & 71.3 \\
        Ours (Trained) & \textbf{59.7} & 88.3 & \textbf{72.6} & \textbf{80.1} & \textbf{73.5} \\
        \bottomrule
    \end{tabular}
    \label{tab:res_planner}
    \vspace{-10pt}
\end{wraptable}

\vspace{-10pt}
\subsection{Ablation Studies} \label{sec:ablation}
\vspace{-6pt}

To rigorously validate our design and training configurations, we conduct a comprehensive series of ablation studies. Specifically, we analyze the impact of Planner optimization and evaluate different Planner reward formulations. We then compare our dynamic approach against a static rubric baseline, and verify the preservation of unimodal capabilities on text-only benchmarks. Furthermore, we assess the necessity of visual context during checklist synthesis. Finally, we demonstrate the framework's generalization across alternative RL algorithms.

\vspace{-6pt}
\paragraph{Ablation on Planner Optimization.}

We investigate the necessity of active rubric optimization by testing a configuration where the Planner is not fine-tuned via RL. In this setting, the Planner operates entirely zero-shot, relying strictly on greedy decoding to generate a static checklist, while the Verifier is trained independently to execute it. As shown in Table \ref{tab:res_planner}, although this frozen configuration provides baseline structural guidance and improves overall accuracy (78.8), removing the Planner's learning signal bottlenecks the framework's capability. Specifically, the untrained Planner struggles to dynamically adapt to complex reasoning, yielding only 66.3 in the Reasoning subcategory. By contrast, our fully trained framework achieves a \textbf{+6.3} point improvement in Reasoning (72.6) compared to the frozen Planner. This demonstrates that multi-task RL is essential for teaching the Planner to synthesize targeted, instance-specific rubrics.

\begin{wraptable}{r}{0.48\textwidth}
    \vspace{-12pt}
    \centering
    \caption{\textbf{Ablation on Planner reward formulation.} VL-RewardBench accuracy. A relative improvement reward ensures proper credit assignment, driving \textbf{+2.5} Overall and \textbf{+3.5} Reasoning gains over an unbaselined absolute reward.}
    \vspace{-5pt}
    \footnotesize 
    % --- THE FIX: Lowered padding to 4pt to respect the margin ---
    \setlength{\tabcolsep}{5pt} 
    \begin{tabular}{lccccc} 
        \toprule
        \textbf{Models} & \textbf{Gen.} & \textbf{Hallu.} & \textbf{Reas.} & \textbf{All} & \textbf{Macro} \\
        \midrule
        Base 8B & 47.0 & 72.4 & 43.2 & 61.3 & 54.2 \\
        Ours (Abs.) & 58.6 & 85.7 & 69.1 & 77.6 & 71.1 \\ 
        Ours (Rel.) & \textbf{59.7} & \textbf{88.3} & \textbf{72.6} & \textbf{80.1} & \textbf{73.5} \\
        \bottomrule
    \end{tabular}
    \label{tab:res_reward}
    \vspace{-12pt}
\end{wraptable}

\vspace{-6pt}
\paragraph{Ablation on Planner Reward Formulation.} 

To validate our choice of the relative improvement reward for the Planner (Eq. \ref{eq: planner_reward}), we evaluate an alternative formulation where the Planner is rewarded strictly based on the absolute accuracy of the final verdict ($R_{plan}(c_i) = \mathbb{I}(z_i = z^*)$). As shown in Table \ref{tab:res_reward}, this formulation degrades performance across all metrics compared to our relative reward. Most notably, the Overall accuracy drops from 80.1 to 77.6 and the Reasoning score reduces from 72.6 to 69.1. This degradation shows the flaw of the unbaselined approach: it fails to account for instances where the model can easily predict the ground truth without any structural guidance. Consequently, the Planner receives positive reinforcement even for generating irrelevant checklists on trivial tasks, leading to noisy gradient updates. This confirms that the Planner should be explicitly incentivized to correct the Verifier's blind spots, rather than merely sharing credit for easy successes.

\begin{wraptable}{r}{0.48\textwidth}
    \vspace{-13pt}
    \centering
    \caption{\textbf{Ablation with a static rubric baseline.} Performance on the VL-RewardBench. While a static rubric provides a useful structural prior, dynamically generating instance-specific checklists drives a large \textbf{+13.0} point gain in the Reasoning subcategory for the 8B model.}
    \vspace{-5pt}
    \footnotesize 
    \setlength{\tabcolsep}{5pt}
    \begin{tabular}{lccccc}
        \toprule
        \textbf{Models} & \textbf{Gen.} & \textbf{Hallu.} & \textbf{Reas.} & \textbf{All} & \textbf{Macro} \\
        \midrule
        Base 4B & 46.4 & 64.9 & 36.0 & 54.9 & 49.1 \\
        No rubric & 51.9 & 87.1 & 50.8 & 72.8 & 63.3 \\
        Static rubric & 47.5 & 85.7 & 62.8 & 74.3 & 65.3 \\
        Ours & \textbf{55.3} & \textbf{87.7} & \textbf{65.9} & \textbf{77.5} & \textbf{69.6} \\
        
        \midrule
        Base 8B & 47.0 & 72.4 & 43.2 & 61.3 & 54.2 \\
        No rubric & 55.8 & 86.1 & 48.3 & 72.1 & 63.4 \\
        Static rubric & 53.6 & 87.2 & 59.6 & 75.3 & 66.8 \\
        Ours & \textbf{59.7} & \textbf{88.3} & \textbf{72.6} & \textbf{80.1} & \textbf{73.5} \\
        \bottomrule
    \end{tabular}
    \label{tab:res_static}
    \vspace{-10pt}
\end{wraptable}

\vspace{-6pt}
\paragraph{Comparison with a Static Rubric Baseline.}

To isolate the value of generating instance-specific checklists, we evaluate a static-rubric baseline. In this setting, the RL-finetuned Verifier is conditioned on a generic, dataset-level evaluation prompt (e.g., “check for hallucinations, logical consistency, and correct object identification”) rather than a dynamically generated checklist. Please see the static-rubric prompt template in Appendix. As shown in Table \ref{tab:res_static}, providing a generic structural prior does offer an improvement over the no-rubric baseline (raising 8B Overall accuracy from 72.0 to 75.3). However, a static prompt fundamentally lacks the capacity to adapt to diverse, instance-specific visual contradictions. Consequently, our dynamic framework (\dr) largely outperforms the static baseline across all metrics. Most notably, dynamically generating the rubric yields a \textbf{+13.0} point gain in the Reasoning subcategory for the 8B model. This confirms that actively synthesize a targeted checklist to guide evaluation is beneficial.

\vspace{-6pt}
\paragraph{Evaluation on Text-Only Benchmark.}

% To ensure that multimodal fine-tuning does not degrade pre-existing language capabilities (i.e., catastrophic forgetting), we evaluate our models on the text-only RewardBench \citep{lambert2025rewardbench}. Furthermore, isolating the text modality provides a critical control experiment for modality dependency. It allows us to verify whether the baseline models' lower performance on multimodal tasks stem from a reasoning deficit or from a modality gap, specifically, the tendency to default to textual priors rather than actively verifying complex visual evidence. Table \ref{tab:res_rb} details these evaluation results. The base Qwen3-VL-8B Instruct model exhibits a strong foundation, achieving an overall text-only accuracy of 81.4. Contrasting this robust unimodal performance with its relatively low multimodal performance (Table \ref{tab:res_vl}) provides empirical confirmation of the modality gap: the base model possesses strong logical capabilities, but fails to apply them when rigorous visual verification is required.

To evaluate whether multimodal fine-tuning degrades pre-existing language capabilities, we additionally benchmark our models on the text-only RewardBench~\citep{lambert2025rewardbench}. This setting also serves as a control experiment for modality dependency, allowing us to verify whether the baseline models' lower performance on multimodal tasks stem from a reasoning deficit or from a modality gap. As shown in Table~\ref{tab:res_rb}, the base Qwen3-VL-8B Instruct model exhibits strong text-only performance with an overall accuracy of 81.4, despite relatively low multimodal performance (Table~\ref{tab:res_vl}). This contrast suggests a clear modality gap: the model possesses strong reasoning ability, but struggles to apply it when explicit visual verification is required.

\begin{wraptable}{r}{0.48\textwidth}
    \vspace{-13pt}
    \centering
    \caption{\textbf{Text-only RewardBench accuracy.} \dr preserves foundational language capabilities while actively improving structural reasoning (\textbf{+6.7}) over the base model.}
    \vspace{-5pt}
    \footnotesize 
    \setlength{\tabcolsep}{4pt}
    \begin{tabular}{lcccccc}
        \toprule
        \textbf{Models} & \textbf{Chat} & \textbf{Hard} & \textbf{Safe} & \textbf{Reas.} & \textbf{All} & \textbf{Macro} \\
        \midrule
        Base 8B & 90.8 & 70.8 & 86.8 & 79.0 & 81.4 & 82.1 \\
        No rubric                     & 89.9 & 68.4 & 87.8 & 80.5 & 81.9 & 82.0 \\
        Ours                          & \textbf{91.3} & \textbf{71.9} & \textbf{89.1} & \textbf{85.7} & \textbf{84.6} & \textbf{83.9} \\
        \bottomrule
    \end{tabular}
    \label{tab:res_rb}
    \vspace{-10pt}
\end{wraptable}

% Crucially,  multimodal finetuning with \dr not only prevents language capability degradation but actively enhances foundational logic. Our approach elevates the base model's overall text-only accuracy to 84.6. The most notable improvement occurs in the Reasoning subcategory (from 79.0 to 85.7). This indicates that explicitly generating a disagreement checklist forces the model to decompose complex logical evaluations into sequential, verifiable steps, improving structural reasoning even in the absence of images. Ultimately, while \dr improves text-only performance, its multimodal gains are significantly larger (Table \ref{tab:res_vl}). This contrast confirms that \dr does more than just boost general reasoning; it specifically enforces visual verification to bridge the modality gap.

Crucially, multimodal fine-tuning with \dr not only avoids catastrophic forgetting, but further improves text-only reasoning performance, increasing overall accuracy to 84.6. The largest gain appears in the Reasoning subset, improving from 79.0 to 85.7. These results suggest that disagreement-driven checklist generation encourages the model to decompose evaluations into structured, verifiable reasoning steps, benefiting reasoning even without visual input. At the same time, the substantially larger gains on multimodal benchmarks indicate that \dr specifically improves grounded visual verification rather than merely enhancing general reasoning ability.

\vspace{-6pt}
\paragraph{Text-Only Planning.}

\begin{wraptable}{r}{0.48\textwidth}
    \vspace{-12pt}
    \centering
    \caption{\textbf{Ablation on visual context for the Planner.} Accuracy on VL-RewardBench. While generating a checklist without the image provides strong structural guidance, providing full visual context is helpful for complex visual reasoning tasks.}
    \vspace{-5pt}
    \footnotesize 
    \setlength{\tabcolsep}{4pt}
    \begin{tabular}{lccccc}
        \toprule
        \textbf{Models} & \textbf{Gen.} & \textbf{Hallu.} & \textbf{Reas.} & \textbf{All} & \textbf{Macro} \\
        \midrule
        Base 8B & 47.0 & 72.4 & 43.2 & 61.3 & 54.2 \\
        Ours (Text-Only) & \textbf{61.4} & 88.0 & 68.0 & 79.1 & 72.5 \\ 
        Ours (Full) & 59.7 & \textbf{88.3} & \textbf{72.6} & \textbf{80.1} & \textbf{73.5} \\
        \bottomrule
    \end{tabular}
    \label{tab:res_text_plan}
    \vspace{-12pt}
\end{wraptable}

We conduct another ablation to determine whether the Planner requires the reference image to generate effective checklists. In this text-only Planner setting, the model is provided solely with the two candidate responses to identify disagreements, without seeing the image. As shown in Table \ref{tab:res_text_plan}, while this purely linguistic extraction yields a solid improvement over the 8B base model, it is slightly worse on average than the full \dr approach, especially for the Reasoning subcategory. This demonstrates that while identifying textual contradictions provides a strong structural prior, seeing the image is helpful for the Planner to generate targeted, context-aware constraints that successfully resolve complex visual reasoning tasks.

\vspace{-6pt}
\paragraph{Generalization to Alternative RL Algorithm.} 
While our experiments utilize GRPO, the \dr framework is agnostic to the underlying RL algorithm. To demonstrate this, we evaluate our approach using an alternative algorithm: Decoupled Clip and Dynamic Sampling Policy Optimization (DAPO) \citep{yu2025dapo}. Because our decoupled advantage estimation strictly isolates the reward signals for the Planner and the Verifier, DAPO can be seamlessly integrated to govern the joint policy updates. As shown in Table \ref{tab:res_dapo} (Appendix \ref{app:dapo}) of evaluation on VL-RewardBench, optimizing \dr with DAPO yields solid gains. It outperforms both the no-rubric and static-rubric baselines, improving the Overall accuracy of the 4B and 8B base models by 21.0 and 16.6 points, respectively. This reaffirms the effectiveness of our approach.

\vspace{-8pt}
\section{Conclusions} \label{sec: con}
\vspace{-8pt}

In this work, we introduce DeltaRubric, an approach that reframes multimodal evaluation as an active, two-step visual investigation. By decomposing evaluation into a \textit{Disagreement Planner} and a \textit{Checklist Verifier}, the model is encouraged to isolate factual contradictions and ground its judgments in visual evidence rather than relying on textual priors. This design mitigates lazy judging and improves evaluation reliability. Optimized via multi-role reinforcement learning, \dr substantially outperforms both no-rubric and static-rubric baselines. Moreover, results on the text-only RewardBench show that multimodal fine-tuning with \dr not only avoids catastrophic forgetting of language capabilities, but also improves core reasoning performance. Overall, our findings suggest that decomposing evaluation into structured, verifiable steps leads to more reliable and generalizable multimodal reward modeling. For a discussion on future work, please see Appendix~\ref{sec: limit}.

%%%%%%%%%%%%%%%%%%%%%%%%%%%%%%%%%%%%%%%%%%%%%%%%%%%%%%%%%%%%

\bibliography{ref}

@article{wei2022chain,
  title={Chain-of-thought prompting elicits reasoning in large language models},
  author={Wei, Jason and Wang, Xuezhi and Schuurmans, Dale and Bosma, Maarten and Xia, Fei and Chi, Ed and Le, Quoc V and Zhou, Denny and others},
  journal={Advances in neural information processing systems},
  volume={35},
  pages={24824--24837},
  year={2022}
}

@inproceedings{yu2025rlaif,
  title={Rlaif-v: Open-source ai feedback leads to super gpt-4v trustworthiness},
  author={Yu, Tianyu and Zhang, Haoye and Li, Qiming and Xu, Qixin and Yao, Yuan and Chen, Da and Lu, Xiaoman and Cui, Ganqu and Dang, Yunkai and He, Taiwen and others},
  booktitle={Proceedings of the Computer Vision and Pattern Recognition Conference},
  pages={19985--19995},
  year={2025}
}

@article{wang2025unified,
  title={Unified multimodal chain-of-thought reward model through reinforcement fine-tuning},
  author={Wang, Yibin and Li, Zhimin and Zang, Yuhang and Wang, Chunyu and Lu, Qinglin and Jin, Cheng and Wang, Jiaqi},
  journal={arXiv preprint arXiv:2505.03318},
  year={2025}
}

@article{zhang2025r1,
  title={R1-reward: Training multimodal reward model through stable reinforcement learning},
  author={Zhang, Yi-Fan and Lu, Xingyu and Hu, Xiao and Fu, Chaoyou and Wen, Bin and Zhang, Tianke and Liu, Changyi and Jiang, Kaiyu and Chen, Kaibing and Tang, Kaiyu and others},
  journal={arXiv preprint arXiv:2505.02835},
  year={2025}
}

@article{wang2026msrl,
  title={MSRL: Scaling Generative Multimodal Reward Modeling via Multi-Stage Reinforcement Learning},
  author={Wang, Chenglong and Huo, Yifu and Gan, Yang and He, Qiaozhi and Meng, Qi and Li, Bei and Wang, Yan and Liu, Junfu and Zhou, Tianhua and Zhu, Jingbo and others},
  journal={arXiv preprint arXiv:2603.25108},
  year={2026}
}

@misc{zheng2025easyr1,
  title        = {EasyR1: An Efficient, Scalable, Multi-Modality RL Training Framework},
  author       = {Yaowei Zheng and Junting Lu and Shenzhi Wang and Zhangchi Feng and Dongdong Kuang and Yuwen Xiong and Richong Zhang},
  howpublished = {\url{https://github.com/hiyouga/EasyR1}},
  year         = {2025}
}

@article{zhang2025mm,
  title={Mm-rlhf: The next step forward in multimodal llm alignment},
  author={Zhang, Yi-Fan and Yu, Tao and Tian, Haochen and Fu, Chaoyou and Li, Peiyan and Zeng, Jianshu and Xie, Wulin and Shi, Yang and Zhang, Huanyu and Wu, Junkang and others},
  journal={arXiv preprint arXiv:2502.10391},
  year={2025}
}

@article{bai2025qwen3,
  title={Qwen3-vl technical report},
  author={Bai, Shuai and Cai, Yuxuan and Chen, Ruizhe and Chen, Keqin and Chen, Xionghui and Cheng, Zesen and Deng, Lianghao and Ding, Wei and Gao, Chang and Ge, Chunjiang and others},
  journal={arXiv preprint arXiv:2511.21631},
  year={2025}
}

@inproceedings{li2025vl,
  title={VL-RewardBench: a challenging benchmark for vision-language generative reward models},
  author={Li, Lei and Wei, Yuancheng and Xie, Zhihui and Yang, Xuqing and Song, Yifan and Wang, Peiyi and An, Chenxin and Liu, Tianyu and Li, Sujian and Lin, Bill Yuchen and others},
  booktitle={Proceedings of the Computer Vision and Pattern Recognition Conference},
  pages={24657--24668},
  year={2025}
}

@inproceedings{lambert2025rewardbench,
  title={Rewardbench: Evaluating reward models for language modeling},
  author={Lambert, Nathan and Pyatkin, Valentina and Morrison, Jacob and Miranda, Lester James Validad and Lin, Bill Yuchen and Chandu, Khyathi and Dziri, Nouha and Kumar, Sachin and Zick, Tom and Choi, Yejin and others},
  booktitle={Findings of the Association for Computational Linguistics: NAACL 2025},
  pages={1755--1797},
  year={2025}
}

@article{yasunaga2025multimodal,
  title={Multimodal rewardbench: Holistic evaluation of reward models for vision language models},
  author={Yasunaga, Michihiro and Zettlemoyer, Luke and Ghazvininejad, Marjan},
  journal={arXiv preprint arXiv:2502.14191},
  year={2025}
}

@inproceedings{
loshchilov2018decoupled,
title={Decoupled Weight Decay Regularization},
author={Ilya Loshchilov and Frank Hutter},
booktitle={International Conference on Learning Representations},
year={2019},
url={https://openreview.net/forum?id=Bkg6RiCqY7},
}

@inproceedings{xiong2025llava,
  title={Llava-critic: Learning to evaluate multimodal models},
  author={Xiong, Tianyi and Wang, Xiyao and Guo, Dong and Ye, Qinghao and Fan, Haoqi and Gu, Quanquan and Huang, Heng and Li, Chunyuan},
  booktitle={Proceedings of the Computer Vision and Pattern Recognition Conference},
  pages={13618--13628},
  year={2025}
}

@article{zheng2023judging,
  title={Judging llm-as-a-judge with mt-bench and chatbot arena},
  author={Zheng, Lianmin and Chiang, Wei-Lin and Sheng, Ying and Zhuang, Siyuan and Wu, Zhanghao and Zhuang, Yonghao and Lin, Zi and Li, Zhuohan and Li, Dacheng and Xing, Eric and others},
  journal={Advances in neural information processing systems},
  volume={36},
  pages={46595--46623},
  year={2023}
}

@article{ouyang2022training,
  title={Training language models to follow instructions with human feedback},
  author={Ouyang, Long and Wu, Jeffrey and Jiang, Xu and Almeida, Diogo and Wainwright, Carroll and Mishkin, Pamela and Zhang, Chong and Agarwal, Sandhini and Slama, Katarina and Ray, Alex and others},
  journal={Advances in neural information processing systems},
  volume={35},
  pages={27730--27744},
  year={2022}
}

@article{bai2022training,
  title={Training a helpful and harmless assistant with reinforcement learning from human feedback},
  author={Bai, Yuntao and Jones, Andy and Ndousse, Kamal and Askell, Amanda and Chen, Anna and DasSarma, Nova and Drain, Dawn and Fort, Stanislav and Ganguli, Deep and Henighan, Tom and others},
  journal={arXiv preprint arXiv:2204.05862},
  year={2022}
}

@article{zhong2025comprehensive,
  title={A comprehensive survey of reward models: Taxonomy, applications, challenges, and future},
  author={Zhong, Jialun and Shen, Wei and Li, Yanzeng and Gao, Songyang and Lu, Hua and Chen, Yicheng and Zhang, Yang and Zhou, Wei and Gu, Jinjie and Zou, Lei},
  journal={arXiv preprint arXiv:2504.12328},
  year={2025}
}

@article{lambert2025reinforcement,
  title={Reinforcement learning from human feedback},
  author={Lambert, Nathan},
  journal={arXiv preprint arXiv:2504.12501},
  year={2025}
}

@article{rafailov2023direct,
  title={Direct preference optimization: Your language model is secretly a reward model},
  author={Rafailov, Rafael and Sharma, Archit and Mitchell, Eric and Manning, Christopher D and Ermon, Stefano and Finn, Chelsea},
  journal={Advances in neural information processing systems},
  volume={36},
  pages={53728--53741},
  year={2023}
}

@article{saunders2022self,
  title={Self-critiquing models for assisting human evaluators},
  author={Saunders, William and Yeh, Catherine and Wu, Jeff and Bills, Steven and Ouyang, Long and Ward, Jonathan and Leike, Jan},
  journal={arXiv preprint arXiv:2206.05802},
  year={2022}
}

@inproceedings{kim2023prometheus,
  title={Prometheus: Inducing fine-grained evaluation capability in language models},
  author={Kim, Seungone and Shin, Jamin and Cho, Yejin and Jang, Joel and Longpre, Shayne and Lee, Hwaran and Yun, Sangdoo and Shin, Seongjin and Kim, Sungdong and Thorne, James and others},
  booktitle={The Twelfth International Conference on Learning Representations},
  year={2023}
}

@inproceedings{sun2024aligning,
  title={Aligning large multimodal models with factually augmented rlhf},
  author={Sun, Zhiqing and Shen, Sheng and Cao, Shengcao and Liu, Haotian and Li, Chunyuan and Shen, Yikang and Gan, Chuang and Gui, Liangyan and Wang, Yu-Xiong and Yang, Yiming and others},
  booktitle={Findings of the Association for Computational Linguistics: ACL 2024},
  pages={13088--13110},
  year={2024}
}

@inproceedings{yu2024rlhf,
  title={Rlhf-v: Towards trustworthy mllms via behavior alignment from fine-grained correctional human feedback},
  author={Yu, Tianyu and Yao, Yuan and Zhang, Haoye and He, Taiwen and Han, Yifeng and Cui, Ganqu and Hu, Jinyi and Liu, Zhiyuan and Zheng, Hai-Tao and Sun, Maosong and others},
  booktitle={Proceedings of the IEEE/CVF Conference on Computer Vision and Pattern Recognition},
  pages={13807--13816},
  year={2024}
}

@inproceedings{chen2024mllm,
  title={Mllm-as-a-judge: Assessing multimodal llm-as-a-judge with vision-language benchmark},
  author={Chen, Dongping and Chen, Ruoxi and Zhang, Shilin and Wang, Yaochen and Liu, Yinuo and Zhou, Huichi and Zhang, Qihui and Wan, Yao and Zhou, Pan and Sun, Lichao},
  booktitle={Forty-first International Conference on Machine Learning},
  year={2024}
}

@article{ding2025arm,
  title={ARM-Thinker: Reinforcing Multimodal Generative Reward Models with Agentic Tool Use and Visual Reasoning},
  author={Ding, Shengyuan and Fang, Xinyu and Liu, Ziyu and Zang, Yuhang and Cao, Yuhang and Zhao, Xiangyu and Duan, Haodong and Dong, Xiaoyi and Liang, Jianze and Wang, Bin and others},
  journal={arXiv preprint arXiv:2512.05111},
  year={2025}
}

@article{singhal2023long,
  title={A long way to go: Investigating length correlations in rlhf},
  author={Singhal, Prasann and Goyal, Tanya and Xu, Jiacheng and Durrett, Greg},
  journal={arXiv preprint arXiv:2310.03716},
  year={2023}
}

@inproceedings{huang2024opera,
  title={Opera: Alleviating hallucination in multi-modal large language models via over-trust penalty and retrospection-allocation},
  author={Huang, Qidong and Dong, Xiaoyi and Zhang, Pan and Wang, Bin and He, Conghui and Wang, Jiaqi and Lin, Dahua and Zhang, Weiming and Yu, Nenghai},
  booktitle={Proceedings of the IEEE/CVF Conference on Computer Vision and Pattern Recognition},
  pages={13418--13427},
  year={2024}
}

@article{xu2026alternating,
  title={Alternating Reinforcement Learning for Rubric-Based Reward Modeling in Non-Verifiable LLM Post-Training},
  author={Xu, Ran and Liu, Tianci and Dong, Zihan and Yu, Tony and Hong, Ilgee and Yang, Carl and Zhang, Linjun and Zhao, Tao and Wang, Haoyu},
  journal={arXiv preprint arXiv:2602.01511},
  year={2026}
}

@article{guo2025deepseek,
  title={Deepseek-r1: Incentivizing reasoning capability in llms via reinforcement learning},
  author={Guo, Daya and Yang, Dejian and Zhang, Haowei and Song, Junxiao and Wang, Peiyi and Zhu, Qihao and Xu, Runxin and Zhang, Ruoyu and Ma, Shirong and Bi, Xiao and others},
  journal={arXiv preprint arXiv:2501.12948},
  year={2025}
}

@article{lambert2024tulu,
  title={Tulu 3: Pushing frontiers in open language model post-training},
  author={Lambert, Nathan and Morrison, Jacob and Pyatkin, Valentina and Huang, Shengyi and Ivison, Hamish and Brahman, Faeze and Miranda, Lester James V and Liu, Alisa and Dziri, Nouha and Lyu, Shane and others},
  journal={arXiv preprint arXiv:2411.15124},
  year={2024}
}

@article{gunjal2025rubrics,
  title={Rubrics as rewards: Reinforcement learning beyond verifiable domains},
  author={Gunjal, Anisha and Wang, Anthony and Lau, Elaine and Nath, Vaskar and He, Yunzhong and Liu, Bing and Hendryx, Sean},
  journal={arXiv preprint arXiv:2507.17746},
  year={2025}
}

@article{huang2025reinforcement,
  title={Reinforcement learning with rubric anchors},
  author={Huang, Zenan and Zhuang, Yihong and Lu, Guoshan and Qin, Zeyu and Xu, Haokai and Zhao, Tianyu and Peng, Ru and Hu, Jiaqi and Shen, Zhanming and Hu, Xiaomeng and others},
  journal={arXiv preprint arXiv:2508.12790},
  year={2025}
}

@article{liu2025openrubrics,
  title={Openrubrics: Towards scalable synthetic rubric generation for reward modeling and llm alignment},
  author={Liu, Tianci and Xu, Ran and Yu, Tony and Hong, Ilgee and Yang, Carl and Zhao, Tuo and Wang, Haoyu},
  journal={arXiv preprint arXiv:2510.07743},
  year={2025}
}

@article{shao2025dr,
  title={Dr tulu: Reinforcement learning with evolving rubrics for deep research},
  author={Shao, Rulin and Asai, Akari and Shen, Shannon Zejiang and Ivison, Hamish and Kishore, Varsha and Zhuo, Jingming and Zhao, Xinran and Park, Molly and Finlayson, Samuel G and Sontag, David and others},
  journal={arXiv preprint arXiv:2511.19399},
  year={2025}
}

@article{viswanathan2025checklists,
  title={Checklists are better than reward models for aligning language models},
  author={Viswanathan, Vijay and Sun, Yanchao and Ma, Shuang and Kong, Xiang and Cao, Meng and Neubig, Graham and Wu, Tongshuang},
  journal={arXiv preprint arXiv:2507.18624},
  year={2025}
}

@article{jia2025autorubric,
  title={AutoRubric-R1V: Rubric-Based Generative Rewards for Faithful Multimodal Reasoning},
  author={Jia, Mengzhao and Zhang, Zhihan and Cases, Ignacio and Liu, Zheyuan and Jiang, Meng and Qi, Peng},
  journal={arXiv preprint arXiv:2510.14738},
  year={2025}
}

@article{shao2024deepseekmath,
  title={Deepseekmath: Pushing the limits of mathematical reasoning in open language models},
  author={Shao, Zhihong and Wang, Peiyi and Zhu, Qihao and Xu, Runxin and Song, Junxiao and Bi, Xiao and Zhang, Haowei and Zhang, Mingchuan and Li, YK and Wu, Yang and others},
  journal={arXiv preprint arXiv:2402.03300},
  year={2024}
}

@article{sheng2026reinforcing,
  title={Reinforcing Chain-of-Thought Reasoning with Self-Evolving Rubrics},
  author={Sheng, Leheng and Ma, Wenchang and Hong, Ruixin and Wang, Xiang and Zhang, An and Chua, Tat-Seng},
  journal={arXiv preprint arXiv:2602.10885},
  year={2026}
}

@inproceedings{
zhang2026basereward,
title={BaseReward: A Strong Baseline for Multimodal Reward Model},
author={YiFan Zhang and Haihua Yang and Huanyu Zhang and Yang Shi and Zezhou Chen and Haochen Tian and Chaoyou Fu and Kai WU and Bo Cui and Xu Wang and Jianfei Pan and Haotian Wang and Zhang Zhang and Liang Wang},
booktitle={The Fourteenth International Conference on Learning Representations},
year={2026},
url={https://openreview.net/forum?id=EuN5iszF0a}
}

@article{yu2025dapo,
  title={Dapo: An open-source llm reinforcement learning system at scale},
  author={Yu, Qiying and Zhang, Zheng and Zhu, Ruofei and Yuan, Yufeng and Zuo, Xiaochen and Yue, Yu and Dai, Weinan and Fan, Tiantian and Liu, Gaohong and Liu, Lingjun and others},
  journal={arXiv preprint arXiv:2503.14476},
  year={2025}
}

@misc{deepseekai2026deepseekv4,
      title={DeepSeek-V4: Towards Highly Efficient Million-Token Context Intelligence},
      author={DeepSeek-AI},
      year={2026},
}

@inproceedings{trung-etal-2024-reft,
    title = "{R}e{FT}: Reasoning with Reinforced Fine-Tuning",
    author = "Trung, Luong  and
      Zhang, Xinbo  and
      Jie, Zhanming  and
      Sun, Peng  and
      Jin, Xiaoran  and
      Li, Hang",
    editor = "Ku, Lun-Wei  and
      Martins, Andre  and
      Srikumar, Vivek",
    booktitle = "Proceedings of the 62nd Annual Meeting of the Association for Computational Linguistics (Volume 1: Long Papers)",
    month = aug,
    year = "2024",
    address = "Bangkok, Thailand",
    publisher = "Association for Computational Linguistics",
    url = "https://aclanthology.org/2024.acl-long.410/",
    doi = "10.18653/v1/2024.acl-long.410",
    pages = "7601--7614"
}

@article{grattafiori2024llama,
  title={The llama 3 herd of models},
  author={Grattafiori, Aaron and Dubey, Abhimanyu and Jauhri, Abhinav and Pandey, Abhinav and Kadian, Abhishek and Al-Dahle, Ahmad and Letman, Aiesha and Mathur, Akhil and Schelten, Alan and Vaughan, Alex and others},
  journal={arXiv preprint arXiv:2407.21783},
  year={2024}
}

@article{fu2025vita,
  title={Vita-1.5: Towards gpt-4o level real-time vision and speech interaction},
  author={Fu, Chaoyou and Lin, Haojia and Wang, Xiong and Zhang, Yi-Fan and Shen, Yunhang and Liu, Xiaoyu and Cao, Haoyu and Long, Zuwei and Gao, Heting and Li, Ke and others},
  journal={arXiv preprint arXiv:2501.01957},
  year={2025}
}

@article{zhang2024benchmarking,
  title={Benchmarking large multimodal models against common corruptions},
  author={Zhang, Jiawei and Pang, Tianyu and Du, Chao and Ren, Yi and Li, Bo and Lin, Min},
  journal={arXiv preprint arXiv:2401.11943},
  year={2024}
}

@article{zhu2025internvl3,
  title={Internvl3: Exploring advanced training and test-time recipes for open-source multimodal models},
  author={Zhu, Jinguo and Wang, Weiyun and Chen, Zhe and Liu, Zhaoyang and Ye, Shenglong and Gu, Lixin and Tian, Hao and Duan, Yuchen and Su, Weijie and Shao, Jie and others},
  journal={arXiv preprint arXiv:2504.10479},
  year={2025}
}

@inproceedings{chen2024internvl,
    title={Internvl: Scaling up vision foundation models and aligning for generic visual-linguistic tasks},
    author={Chen, Zhe and Wu, Jiannan and Wang, Wenhai and Su, Weijie and Chen, Guo and Xing, Sen and Zhong, Muyan and Zhang, Qinglong and Zhu, Xizhou and Lu, Lewei and others},
    booktitle={Proceedings of the IEEE/CVF Conference on Computer Vision and Pattern Recognition},
    pages={24185--24198},
    year={2024}
  }

@inproceedings{deitke2025molmo,
  title={Molmo and pixmo: Open weights and open data for state-of-the-art vision-language models},
  author={Deitke, Matt and Clark, Christopher and Lee, Sangho and Tripathi, Rohun and Yang, Yue and Park, Jae Sung and Salehi, Mohammadreza and Muennighoff, Niklas and Lo, Kyle and Soldaini, Luca and others},
  booktitle={Proceedings of the Computer Vision and Pattern Recognition Conference},
  pages={91--104},
  year={2025}
}

@article{dai2024nvlm,
  title={Nvlm: Open frontier-class multimodal llms},
  author={Dai, Wenliang and Lee, Nayeon and Wang, Boxin and Yang, Zhuolin and Liu, Zihan and Barker, Jon and Rintamaki, Tuomas and Shoeybi, Mohammad and Catanzaro, Bryan and Ping, Wei},
  journal={arXiv preprint arXiv:2409.11402},
  year={2024}
}

@article{li2025self,
  title={Self-rewarding vision-language model via reasoning decomposition},
  author={Li, Zongxia and Yu, Wenhao and Huang, Chengsong and Liu, Rui and Liang, Zhenwen and Liu, Fuxiao and Che, Jingxi and Yu, Dian and Boyd-Graber, Jordan and Mi, Haitao and others},
  journal={arXiv preprint arXiv:2508.19652},
  year={2025}
}

@article{liu2025vogue,
  title={Vogue: Guiding exploration with visual uncertainty improves multimodal reasoning},
  author={Liu, Rui and Yu, Dian and Zheng, Tong and Dai, Runpeng and Li, Zongxia and Yu, Wenhao and Liang, Zhenwen and Song, Linfeng and Mi, Haitao and Tokekar, Pratap and others},
  journal={arXiv preprint arXiv:2510.01444},
  year={2025}
}

@article{zheng2025parallel,
  title={Parallel-r1: Towards parallel thinking via reinforcement learning},
  author={Zheng, Tong and Zhang, Hongming and Yu, Wenhao and Wang, Xiaoyang and Dai, Runpeng and Liu, Rui and Bao, Huiwen and Huang, Chengsong and Huang, Heng and Yu, Dong},
  journal={arXiv preprint arXiv:2509.07980},
  year={2025}
}

@article{dai2025cde,
  title={Cde: Curiosity-driven exploration for efficient reinforcement learning in large language models},
  author={Dai, Runpeng and Song, Linfeng and Liu, Haolin and Liang, Zhenwen and Yu, Dian and Mi, Haitao and Tu, Zhaopeng and Liu, Rui and Zheng, Tong and Zhu, Hongtu and others},
  journal={arXiv preprint arXiv:2509.09675},
  year={2025}
}

@article{liu2025stable,
  title={Stable and Efficient Single-Rollout RL for Multimodal Reasoning},
  author={Liu, Rui and Yu, Dian and Ke, Lei and Liu, Haolin and Zhou, Yujun and Liang, Zhenwen and Mi, Haitao and Tokekar, Pratap and Yu, Dong},
  journal={arXiv preprint arXiv:2512.18215},
  year={2025}
}
\bibliographystyle{plainnat}

\newpage
\appendix
\section{Appendix} \label{sec:apppendix}

\subsection{Implementation Details} \label{app:imp}
We perform direct RL training on the Qwen3-VL-4B and 8B Instruct \citep{bai2025qwen3} models. For the Planner, we sample $N=5$ candidate checklists per prompt. For the Verifier, we sample $M=5$ reasoning trajectories per prompt. All models are trained for 120 steps using AdamW~\citep{loshchilov2018decoupled} optimizer with a learning rate of $1\times10^{-6}$ and weight decay of $0.01$. We adopt a global batch size of $128$, a rollout batch size of $256$, and generate rollouts with a temperature $1.0$. The implementation builds on the EasyR1 framework \citep{zheng2025easyr1}. 

% \subsection{Training Cost Analysis} \label{app:train_cost}
\subsection{Sensitivity Analysis} \label{app:sen}

We conduct a sensitivity analysis on the guidance bonus coefficient $\lambda$ for the verifier reward defined in Eq. \ref{eq: verifer_reward}. We test values ranging from $\{0.0,0.2,0.4,0.6\}$, where $\lambda=0.0$ represents a baseline with no guidance bonus. Table \ref{tab:main_res_lambda} presents the performance of the Qwen3-VL-8B Instruct model trained via our approach on VL-RewardBench. The results indicate that applying a guidance bonus outperforms the unguided baseline ($\lambda=0.0$), with $\lambda=0.4$ achieving the optimal balance (Overall Accuracy: 80.1, Macro Avg: 73.5). This confirms that an appropriately scaled bonus effectively incentivizes the verifier to adhere to the generated rubric. However, scaling the coefficient further to $\lambda=0.6$ results in a degradation, particularly in the reasoning sub-category. This drop suggests that an overly aggressive guidance penalty overrides the primary preference signal, constraining the model's capacity for broader contextual reasoning required to render a correct final verdict.

\begin{table*}[ht]
    \centering
    \caption{\textbf{Sensitivity analysis of the guidance bonus coefficient ($\lambda$).} Evaluating our \dr-trained Qwen3-VL-8B Instruct model on VL-RewardBench reveals that the optimal setting ($\lambda=0.4$) achieves the highest overall accuracy and macro average. Disabling the bonus ($\lambda=0.0$) reduces visual grounding, while an excessively high coefficient ($\lambda=0.6$) degrades complex reasoning by causing the model to over-optimize for checklist adherence.}
    \resizebox{0.8\textwidth}{!}{
    \begin{tabular}{lccccc}
        \toprule
        \textbf{Coefficients} & \textbf{General} & \textbf{Hallucination} & \textbf{Reasoning} & \textbf{Overall} & \textbf{Macro Avg} \\       
        \midrule
        $\lambda=0.0$ & 55.3 & 87.9 & 68.5 & 78.2 & 70.5 \\
        $\lambda=0.2$ & 54.1 & 86.3 & \textbf{72.9} & 78.2 & 71.1 \\
        \textbf{$\lambda=0.4$} & \textbf{59.7} & \textbf{88.3} & 72.6 & \textbf{80.1} & \textbf{73.5} \\
        $\lambda=0.6$ & 58.6 & 87.3 & 66.3 & 77.8 & 70.7 \\
        \bottomrule
    \end{tabular}
    }
    \label{tab:main_res_lambda}
    \vspace{-10pt}
\end{table*}

\subsection{Generalization to Alternative RL Algorithm.} \label{app:dapo}
While our experiments utilize GRPO, the \dr framework is agnostic to the underlying RL algorithm. To demonstrate this, we evaluate our approach using an alternative algorithm: Decoupled Clip and Dynamic Sampling Policy Optimization (DAPO) \citep{yu2025dapo}. Because our decoupled advantage estimation strictly isolates the reward signals for the Planner and the Verifier, DAPO can be seamlessly integrated to govern the joint policy updates. As shown in Table \ref{tab:res_dapo} of evaluation on VL-RewardBench, optimizing \dr with DAPO yields solid gains, improving the Overall accuracy of the 4B and 8B base models by 10.0 and 16.6 points, respectively. This reaffirms the effectiveness of our approach.

\begin{table*}[ht]
    \centering
    \caption{\textbf{Evaluation with alternative RL algorithm.} Accuracy on the VL-RewardBench when utilizing DAPO \citep{yu2025dapo} as the underlying RL algorithm. \dr maintains strong performance advantages over the no-rubric and static-rubric baselines.}
    \resizebox{\textwidth}{!}{
    \begin{tabular}{lccccc}
        \toprule
        \textbf{Models} & \textbf{General} & \textbf{Hallucination} & \textbf{Reasoning} & \textbf{Overall} & \textbf{Macro Avg} \\
        \midrule
        Qwen3-VL-4B Instruct \citep{bai2025qwen3} & 46.4 & 64.9 & 36.0 & 54.9 & 49.1 \\
        \quad + No rubric & 54.1 & 86.3 & 56.8 & 74.3 & 66.3 \\
        \quad + Static rubric & 50.8 & 86.3 & 64.7 & 75.6 & 67.3 \\
        \quad + \dr & 54.7 & 86.4 & 63.1 & 75.9 & 68.1 \\
        
        \midrule
        Qwen3-VL-8B Instruct \citep{bai2025qwen3} & 47.0 & 72.4 & 43.2 & 61.3 & 54.2 \\
        \quad + No rubric & 51.4 & 84.7 & 63.4 & 74.2 & 66.5 \\
        \quad + Static rubric & 53.6 & 83.9 & 65.3 & 74.7 & 67.6 \\
        \quad + \dr & 55.8 & 87.5 & 68.8 & 77.9 & 70.1 \\
        \bottomrule
    \end{tabular}
    }
    \label{tab:res_dapo}
    % \vspace{-10pt}
\end{table*}

\subsection{Discussions and Future Work} \label{sec: limit}

While \dr establishes a robust, structurally grounded framework for multimodal evaluation, it also presents promising avenues for future exploration. Future work could explore employing dynamic routing to trigger checklist generation only for highly ambiguous cases. Furthermore, extending the \dr framework to temporal modalities, such as video evaluation, offers an exciting direction for scaling instance-specific verification. 

% Finally, investigating iterative planning, where the Verifier can dynamically prompt the Planner for additional clarifying constraints, could further enhance the model's ability to resolve deeply complex visual contradictions.

% Currently, the explicit two-step Planner-Verifier architecture inherently introduces additional inference overhead compared to standard monolithic evaluation. Future work could explore distilling this active investigation process back into a single-pass model, or

\subsection{Qualitative Examples} \label{app:examples}

\begin{figure*}[ht]
    \centering
    \includegraphics[width=\linewidth]{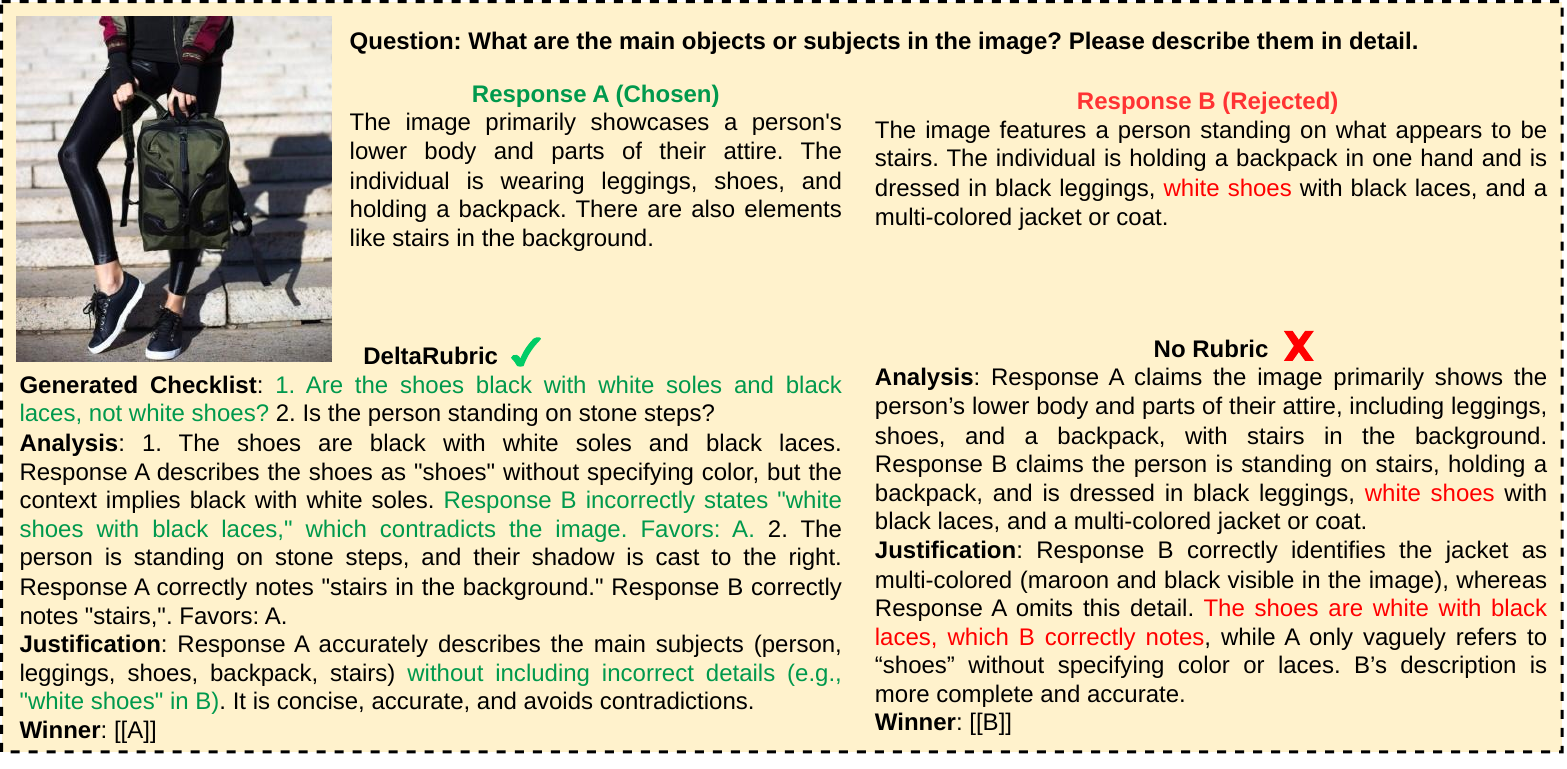}
    \caption{\textbf{Qualitative comparison of evaluation methods.} The standard no-rubric baseline fails to verify fine-grained visual details and incorrectly validates the hallucinated "white shoes" in Response B. In contrast, \dr generates a targeted checklist that explicitly isolates the conflicting shoe color. By enforcing active visual verification, \dr successfully catches the hallucination and correctly selects Response A.}
    \label{fig:case2}
\end{figure*}

\begin{figure*}[ht]
    \centering
    \includegraphics[width=\linewidth]{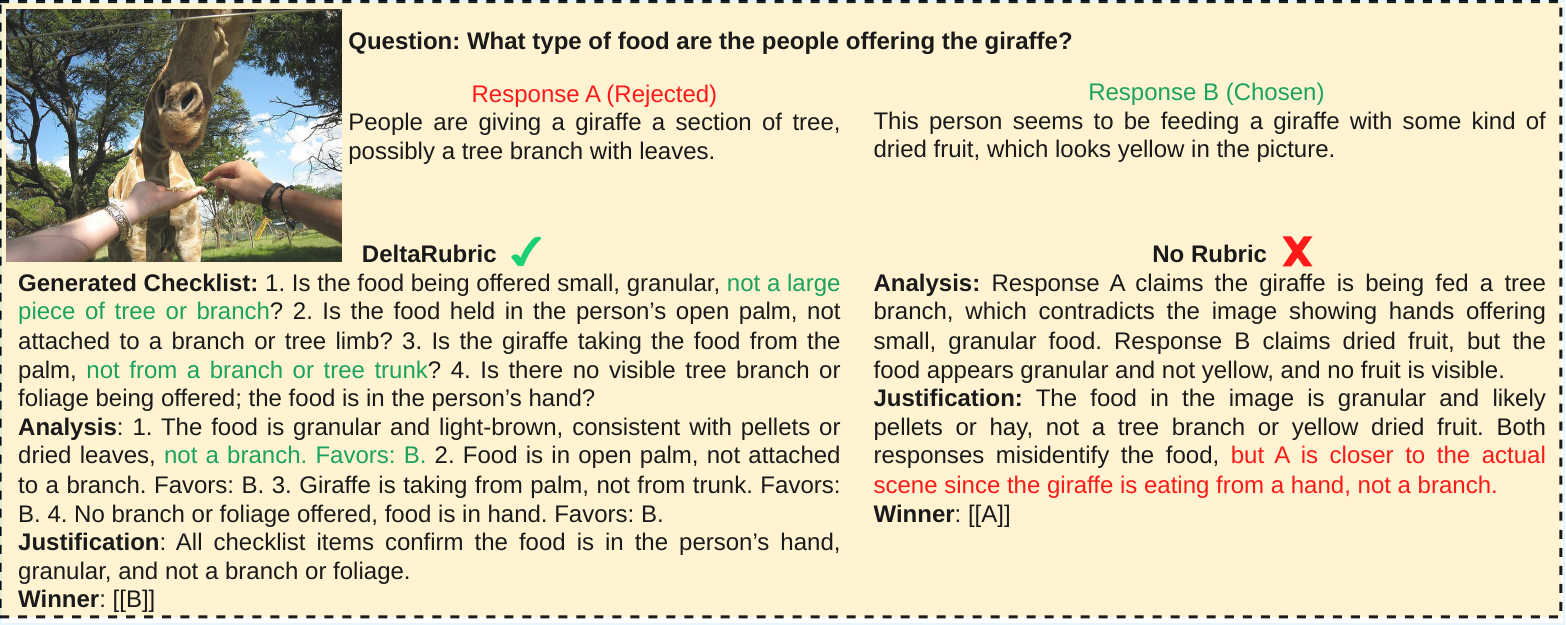}
    \caption{\textbf{Qualitative comparison of evaluation methods.} The standard no-rubric baseline exhibits severe logical inconsistency: it correctly notes the absence of a tree branch but unaccountably still chooses Response A. In contrast, \dr generates a targeted checklist that systematically verifies the visual evidence (e.g., granular food in an open palm, not a branch), strictly enforcing logical consistency and correctly select Response B.}
    \label{fig:case3}
\end{figure*}

\begin{figure*}[ht]
    \centering
    \includegraphics[width=\linewidth]{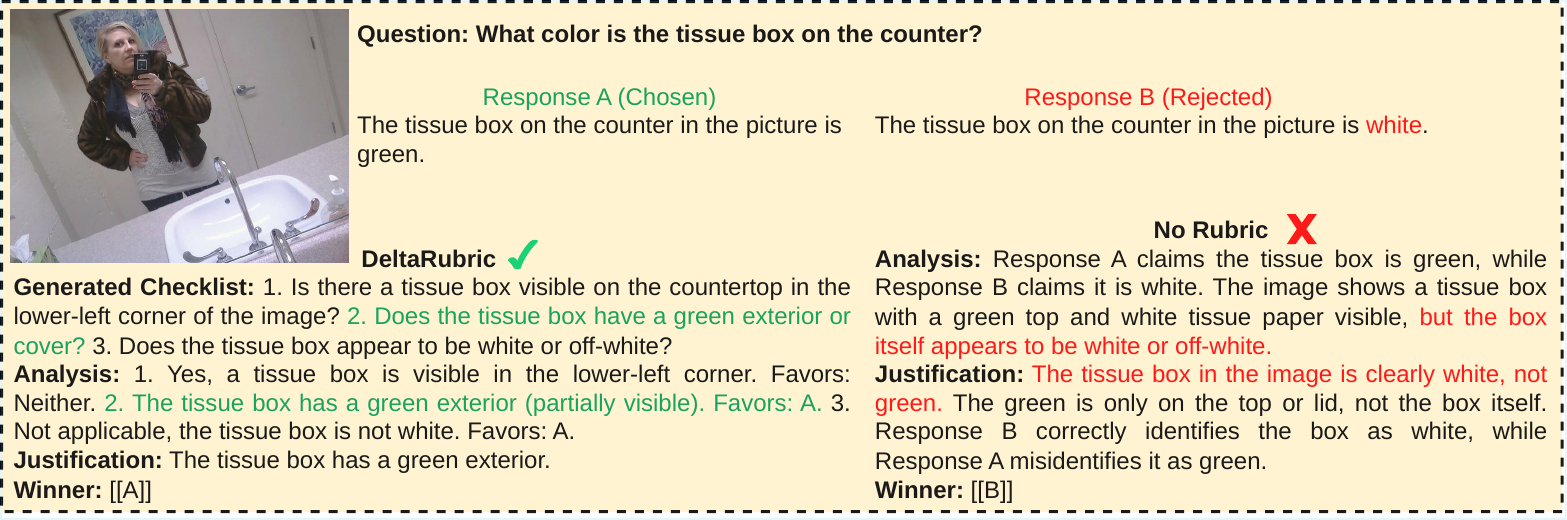}
    \caption{\textbf{Qualitative comparison of evaluation methods.} The standard no-rubric baseline struggles with fine-grained visual attribute binding, confusing the white tissues with the actual color of the box to incorrectly select Response B. In contrast, \dr generates a targeted checklist that explicitly isolates and verifies the "green exterior," preventing this attribute confusion and correctly selecting Response A.}
    \label{fig:case4}
\end{figure*}

We present additional qualitative examples comparing \dr against the standard no-rubric baseline. As illustrated in the following cases (Figure \ref{fig:case2}, \ref{fig:case3} and \ref{fig:case4}), the monolithic baseline frequently suffers from visual hallucinations and flawed reasoning. By contrast, \dr explicitly enforces rigorous visual verification through its generated checklists, systematically mitigating these errors to ensure an accurate final evaluation.

\subsection{Prompt Templates} \label{app:prompt}
Below, we provide the complete set of prompts used to instruct the model to produce structured outputs. This includes the evaluation prompts for the no-rubric and static-rubric baselines, along with the static rubrics, as well as the \dr evaluation prompts. Additionally, we detail the Planner prompts used to generate the checklists and the prompts used for the cheap verdicts.

\begin{tcolorbox}[colback=gray!10, colframe=gray!70, title=No-Rubric Evaluation Prompt]
You are a fair judge. Decide which response better answers the question below based on the image.

Question: \{question\}

Response A: \{response\_a\}

Response B: \{response\_b\}

Compare the two responses and assess which one is better. Then give your overall judgment.

Analysis: <compare both responses>

Justification: <overall reasoning>

Winner: [[A]] or [[B]]
\end{tcolorbox}

\begin{tcolorbox}[colback=gray!10, colframe=gray!70, title=Static-Rubric Evaluation Prompt]
You are a fair judge. Decide which response better answers the question below based on the image.

Question: \{question\}

Evaluation Criteria: \{rubric\}

Response A: \{response\_a\}

Response B: \{response\_b\}

Use the rubric as guidance, not as evidence. If any criterion is irrelevant, too vague, or contradicted by the image/question, ignore that criterion.

For each evaluation criterion, compare the two responses and assess which one is better. Then give your overall judgment.

Analysis:<evaluate criterion by criterion>

Justification: <overall reasoning>

Winner: [[A]] or [[B]]
\end{tcolorbox}

\begin{tcolorbox}[colback=gray!10, colframe=gray!70, title=Static Rubrics]
1. Directly answers the question using the information relevant to the image.

2. Makes factual claims that are consistent with the image and avoids unsupported details.

3. Correctly identifies important visual information when it matters for the question.

4. Uses sound reasoning and logical inference where needed.

5. Gives a clear and complete answer.
\end{tcolorbox}

\begin{tcolorbox}[colback=gray!10, colframe=gray!70, title=\dr Evaluation Prompt]
You are a fair judge. Decide which response better answers the question below based on the image.

Use the verification checklist as a sequence of checks to execute, not as evidence. If a checklist item is irrelevant, too vague, or contradicted by the image/question, ignore that item.

Question: \{question\}

Verification Checklist:\{checklist\}

Response A: \{response\_a\}

Response B: \{response\_b\}

Execute the checklist item by item. For each item, state the evidence and which response it favors. Keep the full answer concise.

Analysis:<work through each checklist item with evidence>

Justification: <one short sentence aggregating the checklist results>

Winner: [[A]] or [[B]]
\end{tcolorbox}

\begin{tcolorbox}[colback=gray!10, colframe=gray!70, title=Planner Prompt]
You are preparing an executable verification checklist for judging which of two responses better answers a visual question. Read both responses, identify the decisive disagreements, and write a short checklist that can be executed item by item.

Question: \{question\}

Response A: \{response\_a\}

Response B: \{response\_b\}

Write a numbered list of 2-4 checks. Rules:

- Each check must describe exactly one concrete fact, relation, or constraint to verify from the image.

- Focus strictly on decisive disagreements or contradictory claims in the responses, not generic advice.

- Keep each check neutral and evidence-seeking.

- Do NOT mention Response A or Response B by name.

- Do NOT say which response is better or correct.

- No preamble, no explanation, only the numbered checks.

Verification Checklist:
\end{tcolorbox}

\begin{tcolorbox}[colback=gray!10, colframe=gray!70, title=No Rubric Cheap Verdict Prompt]
You are a fair judge. Decide which response better answers the question below based on the image.

Question: \{question\}

Response A: \{response\_a\}

Response B: \{response\_b\}

Answer ONLY with [[A]] or [[B]].
\end{tcolorbox}

\begin{tcolorbox}[colback=gray!10, colframe=gray!70, title=Checklist Cheap Verdict Prompt]
You are a fair judge. Decide which response better answers the question below based on the image.

Question: \{question\}

Verification Checklist: \{checklist\}

Response A: \{response\_a\}

Response B: \{response\_b\}

Use the verification checklist only as a shortlist of checks. Answer ONLY with [[A]] or [[B]].
\end{tcolorbox}

\clearpage
\end{document}